\PassOptionsToPackage{unicode}{hyperref}
\PassOptionsToPackage{hyphens}{url}
\PassOptionsToPackage{dvipsnames,svgnames,x11names}{xcolor}
\pdfoutput=1
\documentclass[
  12pt]{article}

\usepackage{amsmath,amssymb}
\usepackage{bm}
\usepackage{iftex}
\usepackage{svg}
\ifPDFTeX
  \usepackage[T1]{fontenc}
  \usepackage[utf8]{inputenc}
  \usepackage{textcomp} 
\else 
  \usepackage{unicode-math}
  \defaultfontfeatures{Scale=MatchLowercase}
  \defaultfontfeatures[\rmfamily]{Ligatures=TeX,Scale=1}
\fi
\usepackage{lmodern}
\usepackage{booktabs}
\usepackage[table]{xcolor}
\usepackage{graphicx}
\definecolor{bestblue}{RGB}{221,236,250}
\newcommand{\best}[1]{\cellcolor{bestblue}$\mathbf{#1}$}
\newcommand{\val}[2]{$#1 \pm #2$}
\ifPDFTeX\else  
\fi
\IfFileExists{upquote.sty}{\usepackage{upquote}}{}
\IfFileExists{microtype.sty}{
  \usepackage[]{microtype}
  \UseMicrotypeSet[protrusion]{basicmath} 
}{}
\makeatletter
\@ifundefined{KOMAClassName}{
  \IfFileExists{parskip.sty}{%
    \usepackage{parskip}
  }{
    \setlength{\parindent}{0pt}
    \setlength{\parskip}{6pt plus 2pt minus 1pt}}
}{
  \KOMAoptions{parskip=half}}
\makeatother
\usepackage{xcolor}
\makeatletter
\ifx\paragraph\undefined\else
  \let\oldparagraph\paragraph
  \renewcommand{\paragraph}{
    \@ifstar
      \xxxParagraphStar
      \xxxParagraphNoStar
  }
  \newcommand{\xxxParagraphStar}[1]{\oldparagraph*{#1}\mbox{}}
  \newcommand{\xxxParagraphNoStar}[1]{\oldparagraph{#1}\mbox{}}
\fi
\ifx\subparagraph\undefined\else
  \let\oldsubparagraph\subparagraph
  \renewcommand{\subparagraph}{
    \@ifstar
      \xxxSubParagraphStar
      \xxxSubParagraphNoStar
  }
  \newcommand{\xxxSubParagraphStar}[1]{\oldsubparagraph*{#1}\mbox{}}
  \newcommand{\xxxSubParagraphNoStar}[1]{\oldsubparagraph{#1}\mbox{}}
\fi
\makeatother

\usepackage{longtable,booktabs,array}
\usepackage{calc} 
\usepackage{etoolbox}
\makeatletter
\patchcmd\longtable{\par}{\if@noskipsec\mbox{}\fi\par}{}{}
\makeatother
\IfFileExists{footnotehyper.sty}{\usepackage{footnotehyper}}{\usepackage{footnote}}
\makesavenoteenv{longtable}
\usepackage{graphicx}
\makeatletter
\def\maxwidth{\ifdim\Gin@nat@width>\linewidth\linewidth\else\Gin@nat@width\fi}
\def\maxheight{\ifdim\Gin@nat@height>\textheight\textheight\else\Gin@nat@height\fi}
\makeatother
\setkeys{Gin}{width=\maxwidth,height=\maxheight,keepaspectratio}
\makeatletter
\def\fps@figure{htbp}
\makeatother

\makeatletter
\@ifpackageloaded{caption}{}{\usepackage{caption}}
\AtBeginDocument{%
\ifdefined\contentsname
  \renewcommand*\contentsname{Table of contents}
\else
  \newcommand\contentsname{Table of contents}
\fi
\ifdefined\listfigurename
  \renewcommand*\listfigurename{List of Figures}
\else
  \newcommand\listfigurename{List of Figures}
\fi
\ifdefined\listtablename
  \renewcommand*\listtablename{List of Tables}
\else
  \newcommand\listtablename{List of Tables}
\fi
\ifdefined\figurename
  \renewcommand*\figurename{Figure}
\else
  \newcommand\figurename{Figure}
\fi
\ifdefined\tablename
  \renewcommand*\tablename{Table}
\else
  \newcommand\tablename{Table}
\fi
}
\@ifpackageloaded{float}{}{\usepackage{float}}
\floatstyle{ruled}
\@ifundefined{c@chapter}{\newfloat{codelisting}{h}{lop}}{\newfloat{codelisting}{h}{lop}[chapter]}
\floatname{codelisting}{Listing}

\makeatother
\makeatletter
\@ifpackageloaded{caption}{}{\usepackage{caption}}
\@ifpackageloaded{subcaption}{}{\usepackage{subcaption}}
\makeatother

\ifLuaTeX
  \usepackage{selnolig}  
\fi
\usepackage[]{natbib}
\usepackage{bookmark}

\IfFileExists{xurl.sty}{\usepackage{xurl}}{} 
\hypersetup{
  pdftitle={Title},
  pdfauthor={Author 1; Author 2},
  pdfkeywords={3 to 6 keywords, that do not appear in the title},
  colorlinks=true,
  linkcolor={blue},
  filecolor={Maroon},
  citecolor={Blue},
  urlcolor={Blue},
  pdfcreator={LaTeX via pandoc}}

\newcommand{\anon}{1}

\usepackage{amsmath, amssymb}
\newcommand{\bx}{\bm{x}}
\newcommand{\bz}{\bm{z}}
\usepackage{amsmath,amssymb}

\newcommand{\RR}{\mathbb{R}}
\usepackage{natbib}

\newcommand{\bc}{\bm{c}}

\usepackage{algorithm}
\usepackage{algorithmic}

\newcommand{\beps}{\bm{\epsilon}}
\newcommand{\bxi}{\bm{\xi}}
\begin{document}

\def\spacingset#1{\renewcommand{\baselinestretch}%
{#1}\small\normalsize} \spacingset{1}


\if1\anon
{
  \title{\bf Generative Bayesian Filtering for State Estimation}
  \author{
    \begin{minipage}{0.95\textwidth}
    \centering
    Lei Cao\textsuperscript{1},
    Sihang Feng\textsuperscript{1},
    Jixin Yan\textsuperscript{3},
    Tao Sun\textsuperscript{1},
    Naichen Shi\textsuperscript{1,2,}\thanks{Corresponding author: \texttt{naichen.shi@northwestern.edu}}\\[0.5em]
    \small
    \textsuperscript{1}Department of Mechanical Engineering\\
    \textsuperscript{2}Department of Industrial Engineering and Management Sciences\\
    \textsuperscript{3}Department of Computer Science\\Northwestern University, Evanston, IL
    \end{minipage}
  }
  \maketitle
}
\fi

\if0\anon
{
  \bigskip
  \bigskip
  \bigskip
  \begin{center}
    {\LARGE\bf Title}
\end{center}
  \medskip
} \fi

\bigskip
\begin{abstract}
The state of a dynamic system evolves over time, switching among several latent modes that govern its observable behavior.
Filtering methods infer the latent state from observations.
Classical filtering approaches, including Kalman filters, typically rely on simple observation models, such as linear-Gaussian models, that are incapable of characterizing the increasingly nonlinear and heterogeneous patterns in high-dimensional sensor signals. 
To tackle the challenge, we propose Generative Bayesian Filtering (GBF), a filtering framework that replaces restrictive observation models with pretrained conditional generative models parametrized by conditional variational autoencoders (CVAE). 
For online inference, GBF performs a Bayesian prediction–update recursion in which the measurement update is formulated as a posterior sampling problem that combines the dynamical prior with the CVAE-induced likelihood. 
The resulting filtering problem is then transformed into a score-based sampling problem, which naturally inherits the flexibility from generative models and the uncertainty quantification capabilities from ensembling. Experiments on synthetic datasets and real-world applications involving manufacturing system monitoring and arrhythmia diagnosis demonstrate that GBF improves state estimation accuracy and robustness relative to baseline approaches.
\end{abstract}

\noindent%
\vfill

\newpage
\spacingset{1.8} 

\section{Introduction}\label{sec-intro}

The Internet of Things (IoT) provides capabilities to connect data streams collected from sensor networks to build industrial world models \citep{shang2026roboscape, yan2025learning, kong2020multi}. 
Sensing signals, including temperature, vibration, pressure, and many more, provide realistic descriptions of the industrial system. Despite the realism, the raw sensing data is often heterogeneous and noisy \citep{kamm2023survey,cheng2023intelligent}. Therefore, a statistical challenge is to derive high-level insights from raw signals for downstream analytics. For example, in manufacturing factories, operators are always interested in estimating the process stability or defect risk from sensor readings \citep{wu2023transformer}. Similarly, in hospitals, the clinical measurements and signals are used to assess patient health conditions \citep{harutyunyan2019multitask} and predict the risk for disease progression. In this paper, we study the problem of latent state estimation from streaming sensing data. \citep{kong2021deep,koganti2019data}.

There are a few statistical challenges in latent state estimation in practical applications \citep{simon2006optimal}. 
First, most observable signals only contain indirect and partial information about system states \citep{surya2024maximum}. Take the health monitoring as an example: the ground truth health condition of a patient is only known after thorough medical examinations that are often costly and time-consuming \citep{o2018overtesting}, while real-time sensors, including heartbeats and blood oxygen, only provide indirect information.
Second, noise and disturbances are ubiquitous in the physical world \citep{zhao2017framework}. Factors including environmental shifts, sensor heterogeneity, tool wear, and cyberattack \citep{teh2020sensor} could all give rise to sensing noise that further complicates the task of system state estimation.

Filtering provides a probabilistic solution to tackle the challenges from partial observation and sensing noise \citep{simon2006optimal} by fusing the historical information and current observation to perform state estimation. The temporal coherence of the latent states encoded by the dynamics models aggregates the information from the streaming sensing data to perform more robust inference. A renowned filtering method is Kalman filtering \citep{khodarahmi2023review},  which uses linear Gaussian models to characterize the evolution of the latent states.  Due to its computational simplicity, Kalman filtering has been widely applied to modern navigation \citep{hasan2009review}, tracking \citep{urrea2021kalman}, and robotics \citep{chen2011kalman} tasks. A caveat, however, is the oversimplified linear Gaussian assumptions in the observation model. This assumption restricts the observation to be a linear function of the latent state and lacks the versatility to model the scenarios where the observation is high-dimensional, while the latent state is expected to be simple. In applications such as robotics visual sensing \citep{bin2025survey}, world model \citep{li2025end}, and clinical monitoring \citep{junaid2025multitask}, more flexible observation models are needed.

Generative AIs provide promising alternative solutions to linear Gaussian models. Deep neural networks pretrained on large-scale datasets have already demonstrated the capability to generate complex, high-dimensional data with impressive quality \citep{ko2023large,xing2024survey,mccarthy2025machine,pmlr-v80-achlioptas18a}. Therefore, a natural question is: can we leverage powerful pretrained generative AI models to augment the flexibility of probability models in filtering tasks?

To address the question, this paper proposes a novel framework called Generative Bayesian Filtering (GBF). The central idea is to leverage conditional generative AI models \citep{sohn2015learning,yoon2026multimodal} to characterize the distribution of observations conditional on system states. Then the observation distribution model is integrated into a posterior sampling framework that combines prior information and current observation. 
Since the generative AI models are usually often differentiable, we use backpropagation to develop score-based sampling techniques to sample from the posterior. By calling the sampling subroutine iteratively, we are able to develop a recursive closed-loop framework for low-dimensional state estimation from high-dimensional data streams. 

GBF harvests multiple benefits from both generative AIs and Bayesian filtering. First, the method only requires pretrained generative models, for which abundant literature provides theoretical justifications \citep{kingma2014auto,khemakhem2020variational,reizinger2022embrace} and practical support \citep{pu2016variational,berahmand2024autoencoders} for a wide range of applications. Our numerical experiments also show that the pretrained generative models demonstrate robust performance for high-dimensional nonlinear observations, even under distribution shifts. Second, the temporal transition model is incorporated through a Bayesian framework, providing adaptivity in modeling latent state transitions. Third, the probabilistic nature of the sampling approach naturally endows the state estimate with uncertainties, which are essential for downstream decision-making.

We summarize the main contributions as follows:
\begin{itemize}
\item We propose GBF, a Bayesian filtering framework for latent state estimation that combines generative AI models with sequential transition probability models for high-dimensional sensor streams. 
    \item In GBF, we design a novel backpropagation scheme to transfer information from high-dimensional observations to the low-dimensional latent state.  We further develop a posterior sampling pipeline based on stochastic gradient Langevin dynamics.    
    
    \item Extensive experiments on synthetic datasets, manufacturing process monitoring, and electrocardiogram (ECG) diagnosis show that GBF improves state-estimation accuracy and robustness relative to baseline filters, especially when the observations are noisy and exhibit distribution shifts.
\end{itemize}

The rest of this paper is organized as follows: 
In section \ref{sec-review}, we review the related work on filtering and generative methods for state estimation. Section \ref{sec-gbf} presents an overview of the proposed methodology and further elaborates on the details of the proposed method. Finally, we evaluate the proposed method on two synthetic datasets in section \ref{sec-num} and two real-world case studies from manufacturing and healthcare in section \ref{sec-case}. The conclusion and future work are discussed in section \ref{sec-conc}.

\section{Related Work}\label{sec-review}

\textbf{Filtering with learned representations.} Filtering is widely applied to time series data, like signal denoising, sensor fusion, and system control \citep{sasiadek2002sensor,goodwin2014adaptive}. 
Classical Kalman filtering and its iterated extensions could handle nonlinear measurements \citep{sibley2006iterated}, but the update process still relies on locally linear assumptions.
Ensemble Kalman filtering provides a scalable approximation for high-dimensional systems \citep{katzfuss2016understanding}, but it still relies on a predefined linear-Gaussian model.
To leverage the power of representation learning methods, extended Kalman VAE \citep{fraccaro2017disentangled,klushyn2021latent} uses a neural network to parametrize the coefficients of the linear model. Furthermore, $K^2$VAE methods \citep{wu2025kvae} adopt the Koopman operator to obtain a system with linear dynamics for efficiency and accuracy. Despite the improved flexibility, the methods above all employ linear Gaussian observation models in a learned latent space. As a result, the update process is usually performed on low-dimensional pseudo-observations, making it hard to capture nonlinear, non-Gaussian, and fine-grained heterogeneous sensor data. 

Nonlinear Kalman filtering methods, like unscented Kalman filters \citep{simon2010kalman,gultekin2017nonlinear}, relax the linearity assumption by sigma-point approximation. However, these methods still typically require an explicit observation function, together with covariance propagation under Gaussian assumptions, which limits the flexibility when the observation process is difficult to specify analytically.
Deep state-space models such as Deep Kalman Filters \citep{krishnan2015deep} and related approaches \citep{karl2017deep} jointly learn latent dynamics and observation models through variational objectives. However, the end-to-end framework couples the observation model, transition model, and inference network, making it difficult to use pretrained generative observation models or physics-informed priors without retraining the full model.
In contrast, our proposed GBF utilizes a modularized observation model, which is then incorporated into the filtering framework.

\textbf{Conditional generative model.} Generative AI models aim to capture the distribution of data \citep{lee2023convergence, ruthotto2021introduction,vahdat2020nvae}. By learning to connect the data distribution with tractable distributions such as Gaussian distributions \citep{song2019generative}, the model could generate new samples that possess similar properties with training data \citep{cui2024analysis}. Variational autoencoders \citep{cemgil2020autoencoding,singh2021overview} is a computationally efficient method of generative models, where an encoder is trained to convert data into latent representations that follow Gaussian distributions and a decoder is trained to reconstruct the original data from latent embeddings. In order to generate specified samples, conditional variational autoencoders \citep{pol2019anomaly,kingma2014semi} is proposed to provide a scalable framework for learning conditional distributions \(p(\bx\mid \bc)\) and are widely used for conditional generation. Here $\bx$ represents data and $\bc$ represents the class of data. CVAE has been applied to multiple generative tasks, including trajectory generation \citep{feng2019vehicle}, dialogue generation~\citep{zhao2017learning}, and text-to-speech generation~\citep{kim2021conditional}. Recent works also study multivariate dependence structures in latent representations \citep{oubari2026multicomponent}. However, these formulations do not explicitly address temporal dependence for online state estimation.

\textbf{Inference using generative models.} 
The inherent statistical property of generative models could be utilized for state inference. For example, the latent embeddings of a well-trained CVAE contain the meaningful information of the data \citep{lin2022uncertainty,tibermacine2025conditional}, so the embeddings could be injected into a neural network for state inference.  In addition to classification tasks, the latent variables could be utilized for unseen fault detection \citep{he2021deep,yong2020bayesian}. Although generative models are  used to extract features, those methods are still discriminative.
As a result, their uncertainty quantification capability is usually limited to softmax-based confidence, which may not be well-calibrated.
Given the observation, the classification task could be formulated as a conditional generative inference task \citep{zimmermann2021score,oraki2026scale}, and then the label with the highest likelihood or lowest energy could be assigned. 
However, the candidate classes are compared based on heuristic likelihood, and this method does not model any temporal dynamics of the system.
Similarly, Deep Bayes \citep{li2019generative} replaces the naive conditional model with a deep latent-variable generative model and performs classification by approximating Bayes’ rule using Monte Carlo sampling of the latent variable. In contrast, GBF optimizes the latent seed and categorical variable simultaneously for more accurate inference. 
The score-based update of GBF is similar to semi-amortized and iterative inference methods \citep{kim2018semi,marino2018iterative} that refine latent variables or variational parameters via gradient-based updates. 
Related work studies connections between iterative inference and EM-type procedures \citep{boutin2020iterative}. Those methods share a similar iterative idea with GBF, but they are primarily designed for generation or reconstruction tasks.
Therefore, none of the existing approaches explicitly incorporate temporal prior information and score-based posterior sampling for state estimation from a data stream.

\section{Generative Bayesian Filtering}\label{sec-gbf}

\begin{figure}
\centering{
\includegraphics[width=4in,height=\textheight]{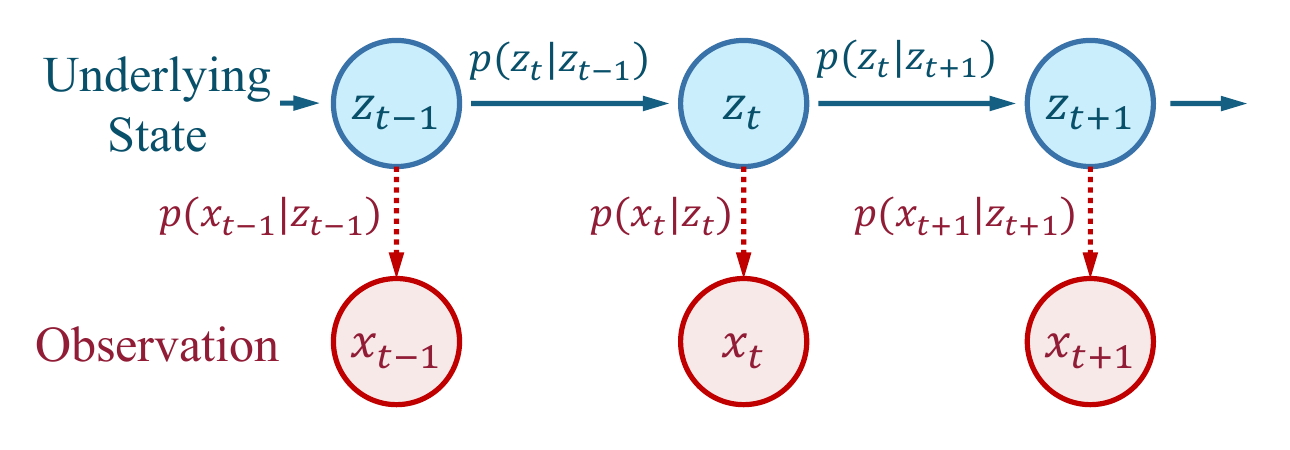}
}
\caption{\label{state_trans} Causal structure of the underlying state and observation of dynamic systems.}
\end{figure}

We briefly describe the mathematical context of probabilistic filtering before introducing GBF. Suppose the underlying system state at time $t$ is $\bz_t\in\RR^{d_z}$, and we can obtain an indirect and high-dimensional observation $\bx_t\in\RR^{d_x} $. We consider the setting where $d_x\gg d_z$, i.e., the observations are replete with details while the states encode aggregated insights. The statistical task is to infer the current state $\bz_t$ from historical measurement $\bx$, that is, estimate the distribution
$
p(\bz_t|\bx_1,\bx_2,\cdots,\bx_t)
$.
We assume a temporal causal structure delineated by the probabilistic model in Figure \ref{state_trans},  which suggests that $\bz_{t-1}$ is a sufficient statistic of $\bx_1,\cdots,\bx_{t-1} $, i.e., $p(\bz_t|\bx_1,\bx_2,\cdots,\bx_t)=p(\bz_t|\bz_{t-1},\bx_t)$.

Given the previous state variable $\bz_{t-1}$ and current sensor measurements $\bx_{t}$, the posterior distribution of the current state could be derived from the Bayesian theorem:
\begin{align}
p(\bz_t \mid \bx_t, \bz_{t-1})
&= \frac{
p(\bz_t, \bx_t, \bz_{t-1})
}{
p(\bx_t, \bz_{t-1})
} \nonumber = \frac{
p(\bx_t \mid \bz_t, \bz_{t-1})
p(\bz_t \mid \bz_{t-1})
p(\bz_{t-1})
}{
p(\bx_t, \bz_{t-1})
} \nonumber \\
&= \frac{
p(\bx_t \mid \bz_t)
p(\bz_t \mid \bz_{t-1})
p(\bz_{t-1})
}{
p(\bx_t, \bz_{t-1})
}  = \frac{
p(\bx_t \mid \bz_t)
p(\bz_t \mid \bz_{t-1})
}{
p(\bx_t \mid \bz_{t-1})
},
\label{eq:posterior_z}
\end{align}
where the third equality comes from the Markov property that $p(\bx_{t}\mid \bz_{t},\bz_{t-1})=p(\bx_{t}\mid \bz_{t})$. From~\eqref{eq:posterior_z}, there are three components in the estimate of posterior distribution: (i) first we need to model the \textit{state transition} $p(\bz_t \mid \bz_{t-1})$, (ii) then calculate the \textit{observation likelihood} $p(\bx_t \mid \bz_t)$, and (iii) handle the denominator $p(\bx_t \mid \bz_{t-1})$. 

In the filtering task, $p(\bz_t\mid \bz_{t-1})$ characterizes the dynamics of the latent state, which is a prior to describe the system evolution dynamics, such as the natural trend in the manufacturing process to move to states associated with defect generation. It is common to use physics-based models, such as PDEs \citep{pagani2017efficient}, or data-driven models, such as auto-regressive models \citep{wang2025larp}, to parametrize the prior information. 

The data integration is achieved through the conditional probability term $p(\bx_{t}\mid \bz_t)$ that describes the likelihood of observing $\bx_t$ given the underlying state $\bz_t$. As discussed before, characterizing this conditional probability is challenging in practice as this distribution is often complex and high-dimensional. 

In~\eqref{eq:posterior_z}, $p(\bx_t \mid \bz_{t-1})$ serves as a normalized constant, which ensures that the posterior distribution $p(\bz_t \mid \bx_t, \bz_{t-1})$ is a valid probability distribution. However, this term is usually analytically intractable, making the precise probabilistic modeling of $p(\bz_t\mid \bx_t,\bz_{t-1})$ challenging. 

\subsection{Overview of GBF}
\begin{figure}
\centering{
\includegraphics[width=4in,height=\textheight]{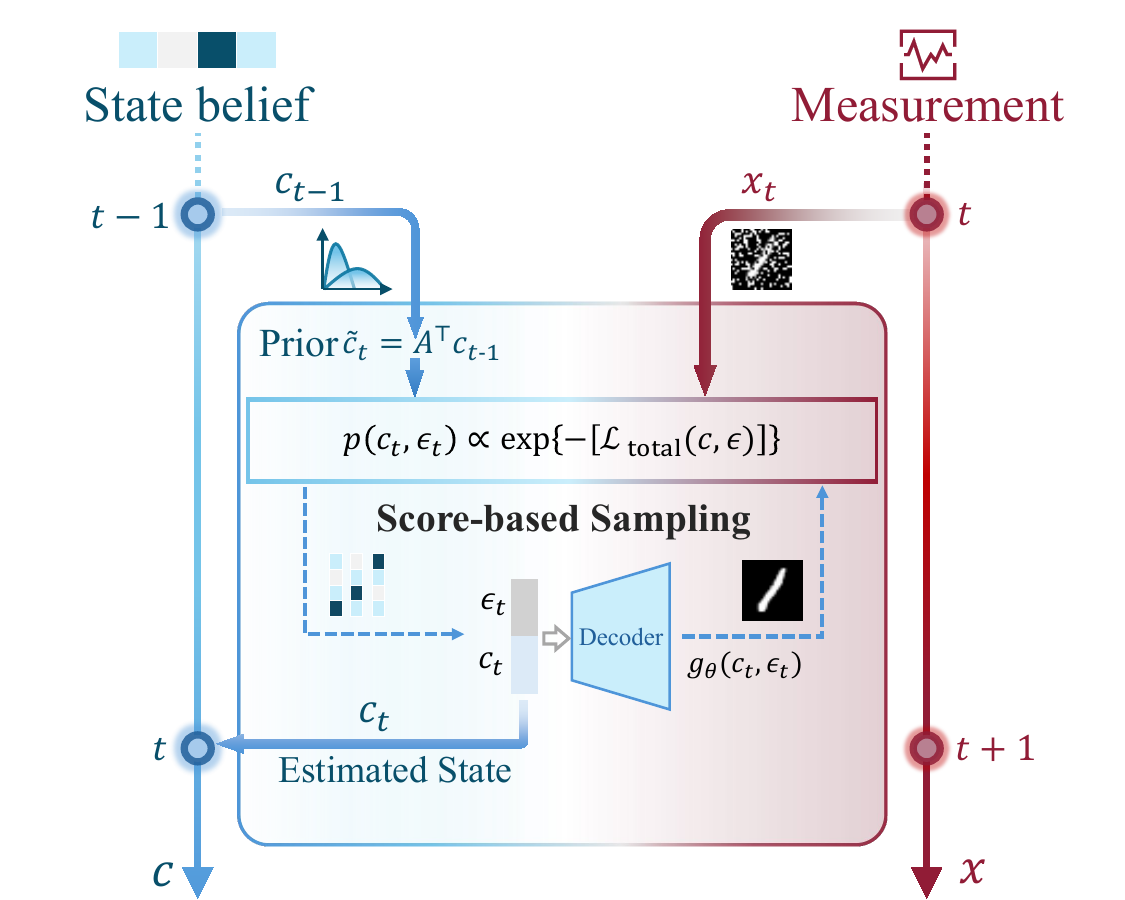}
}
\caption{\label{gbf-overview}Flow diagram of the proposed Generative Bayesian Filtering framework.}
\end{figure}

In light of these challenges, we propose Generative Bayesian Filtering (GBF), a filtering framework that integrates a generative observation model into a recursive inference framework for dynamic systems. The workflow of the proposed GBF framework is illustrated in Figure \ref{gbf-overview}. 

At time step $t$, GBF takes the previous state belief $\bc_{t-1}$ and propagates it through the transition model to obtain the prior belief $\tilde{\bc}_t$. Meanwhile, the noisy and indirect measurement $\bx_t$ is obtained from the sensing stream and used to calibrate the estimated state. To assimilate the new observation, the latent variables are inferred such that the decoder output can match the current measurement while remaining consistent with the dynamical prior. In this way, the framework combines the prediction from the system dynamics with the reconstruction capability of the learned generative model.
In the rest of this section, we will introduce the novel designs in GBF in detail.

\subsection{CVAE for Inference}\label{sec-cave}

A critical design in GBF is the use of Conditional Variational Autoencoders (CVAE)~\citep{kim2021conditional} for probabilistic inference. However, since CVAE was proposed as a generative model, it is not readily apparent how to use it as an inference tool.  This section details our reformulation of the CVAE as a classification model with natural probabilistic interpretations.

\begin{figure}[h]
\centering{
\includegraphics[width=5in,height=\textheight]{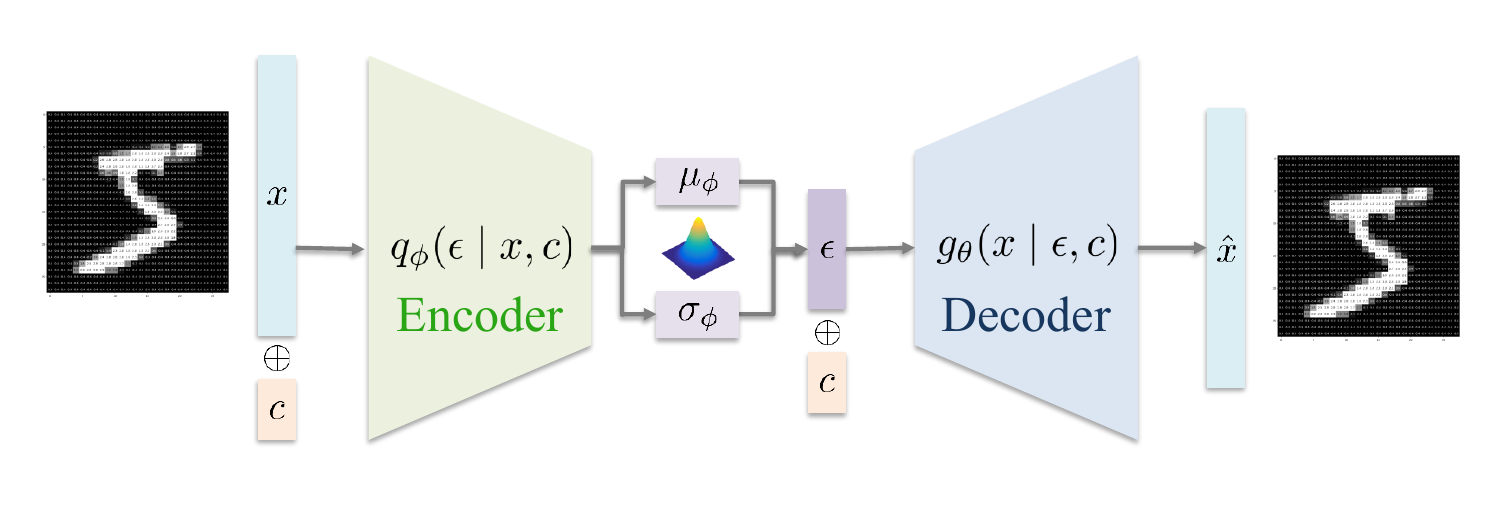}
}
\caption{\label{CVAE}Pipeline of CVAE for classifier.}
\end{figure}

As shown in Figure \ref{CVAE}, CVAE consists of an encoder $q_\phi$ and a decoder $g_\theta$. 
Given an observation $\bx \in \mathbb{R}^{d_x}$ and a categorical class label $\bc$, the encoder, denoted by $q_\phi(\beps\mid \bx,\bc)$, maps the input to a low-dimensional continuous embedding $\beps \in \mathbb{R}^{d_\epsilon}$. 
The decoder then reconstructs the original data $\bx$ from this embedding $\beps$ alongside the conditional label $\bc$. 
We use the complete latent state $\bz$ to denote the concatenation of the class label $\bc$ and the CVAE embedding $\beps$, that is, $\bz = [\bc, \beps]$. 

We leverage an important property of CVAE that the decoder, denoted by $g_{\theta}$, essentially parametrizes a likelihood model. In literature, the distribution $p(\bx \mid \bz)$ is often modeled by a Gaussian distribution $\bx \mid \bz \sim \mathcal{N}(g_{\theta}(\bz),\sigma^2 I)$, where $\sigma$ represents measurement noise, or a Bernoulli distribution $\bx \mid \bz \sim \text{Ber}(g_{\theta}(\bz))$.
Under this probabilistic model, it is natural to reuse the decoder output $g_{\theta}(\bz)$ to parametrize the observation likelihood. Since $g_\theta$ is often implemented as a deep neural network that can act as a universal function approximator~\citep{lu2020universal}, such modeling of $p(\bx \mid \bz)$ is extremely flexible and extends beyond the function class of unscented Kalman filters~\citep{wan2000unscented}.
Therefore, the probabilistic formulation of the decoder is exceptionally well-suited for modeling the observation distribution $p(\bx \mid \bz)$ used in Bayesian filtering.

In this paper, we focus on categorical class variables $\bc$, which is natural to represent the discrete health states or anomaly types of a dynamic system. Hence, the class-related condition $\bc$ is represented as a one-hot encoded vector corresponding to the ground-truth category, and the dimensionality of $\bc$ equals the number of possible system states. 

\textbf{Training:} We briefly review the training process of CVAE. Let $q_{\phi}(\beps \mid \bx, \bc)$ denote the approximate posterior modeled by the encoder, and $g_{\theta}(\bx \mid \beps, \bc)$ denote the likelihood modeled by the decoder. The CVAE is optimized by maximizing the Evidence Lower Bound (ELBO). For a given observation $\bx$ and its corresponding condition $\bc$, the ELBO objective is defined as 
$
\mathcal{L}_{\text{ELBO}}(\bx,\bc)
=  \mathbb{E}_{q_{\phi}(\beps \mid \bx, \bc)}
\big[\log g_{\theta}(\bx \mid \beps, \bc)\big] 
 - D_{\text{KL}}\big(q_{\phi}(\beps \mid \bx, \bc)\,\|\, p(\beps)\big) $,
where $p(\beps)$ is the prior of the latent variable, typically assumed to be a standard isotropic Gaussian $\mathcal{N}(0, I)$. During training, we compute the ELBO over an offline dataset $\mathcal{D} = \{\bc^{(i)},\bx^{(i)}\}_{i=1}^M$ consisting of paired condition-observation tuples. Utilizing the reparameterization trick introduced by Kingma \citep{kingma2013auto}, we jointly optimize the encoder and decoder parameters to maximize the empirical ELBO:
\begin{equation}
    \hat\theta,\hat\phi = \arg\max_{\theta,\phi}\frac{1}{M}\sum_{i=1}^M\mathcal{L}_{\text{ELBO}}(\bx^{(i)},\bc^{(i)}) 
\end{equation}

\textbf{Inference:} Although the trained decoder $g_{\hat\theta}$ is primarily designed to generate new samples from conditional inputs, we demonstrate that it can also function as a robust inference model for classification. 
Given a new, unlabeled observation $\bx$, we transform the classification task into a posterior sampling task. 
Following the probability model $\bx \mid \bz \sim \mathcal{N}(g_{\hat\theta}(\bc,\beps),\sigma^2 I)$, the conditional distribution of the latent variables $(\bc,\beps)$ given the observation $\bx$ can be calculated using the Bayes' law:

\begin{equation}
\label{eqn:vaeclassification}
\pi(\bc,\beps \mid \bx)
\propto
\exp\{-\mathcal{L}_{\mathrm{cls}}(\bc,\beps;\bx)\},\,\text{where}\,
\mathcal{L}_{\mathrm{cls}}=\|\bx - g_{\hat\theta}(\bc,\beps) \|^2+\sigma^2 \|\beps\|^2
\end{equation}
The rationale behind \eqref{eqn:vaeclassification} is intuitive: for a given observation $\bx$, the latent variables that are most likely to have generated $\bx$ with the learned CVAE decoder are favored by the posterior density.
Therefore, the $\bc$ that minimizes $\mathcal{L}_{\mathrm{cls}}$ in~\eqref{eqn:vaeclassification} corresponds to the maximizing a posteriori (MAP) estimate of the class associated with sample $\bx$. 

In practice, to solve the MAP problem, we randomly initialize $N$ samples of $(\bc,\beps)$ from the prior of the encoder, and then leverage a gradient-based optimization method to minimize $\mathcal{L}_{\mathrm{cls}}$. To ensure $\bc$ remains a valid probability distribution during optimization, we parameterize it using logits $\bxi$ and apply a softmax transformation $\bc=\mathrm{softmax}(\bxi)$. The final discrete estimation is obtained by selecting the class corresponding to the highest probability value in the optimized vector $\hat{\bc}$.

\begin{figure}
\centering{
\includegraphics[width=3.5in,height=\textheight]{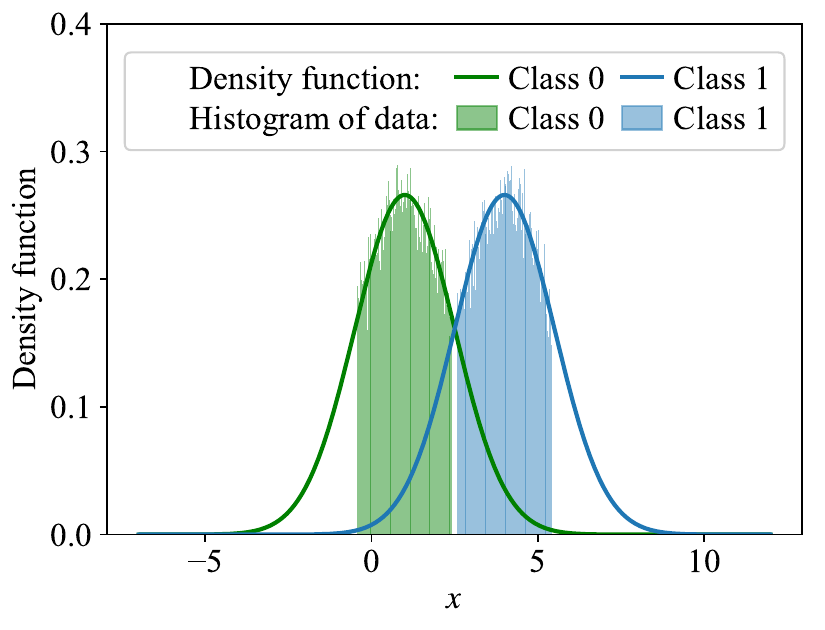}
}
\caption{\label{Gaussian_mixture_data}Data distribution of Gaussian mixture dataset. Solid curves denote the ground truth density functions, while the histogram denotes the censored samples.}
\end{figure}

\textbf{A proof-of-concept experiment on Gaussian mixture inference:} To demonstrate the difference between the proposed CVAE-classifier and the discriminative classifier, we compare the two methods on a synthetic 1D Gaussian mixture inference task. Assume there are two different Gaussian components, where the sampled data is labeled as \textit{Class 0} and \textit{Class 1}. We further assume that the training data is censored, thus concentrated in high-probability regions of the individual Gaussian component, as shown in Figure \ref{Gaussian_mixture_data}.

Define the probability density functions of two Gaussian distributions as $P(x\mid y=0)$ and $P(x\mid y=1)$, then the ground-truth posterior probability for a given $x$ is
\begin{equation}
\label{eqn:gaussianmix}
\begin{aligned}
P_{\mathrm{TRUE}}(y=0 \mid x)
&= \frac{p(x\mid y=0)}
{p(x\mid y=0)+p(x\mid y=1)}, \\
P_{\mathrm{TRUE}}(y=1 \mid x)
&= \frac{p(x\mid y=1)}
{p(x\mid y=0)+p(x\mid y=1)} .
\end{aligned}
\end{equation}
The inference task aims to build statistical estimates that approximate $P_{\mathrm{TRUE}}(y \mid x)$ as well as possible.

\begin{figure}[h]
\centering{
\includegraphics[width=3.5in,height=\textheight]{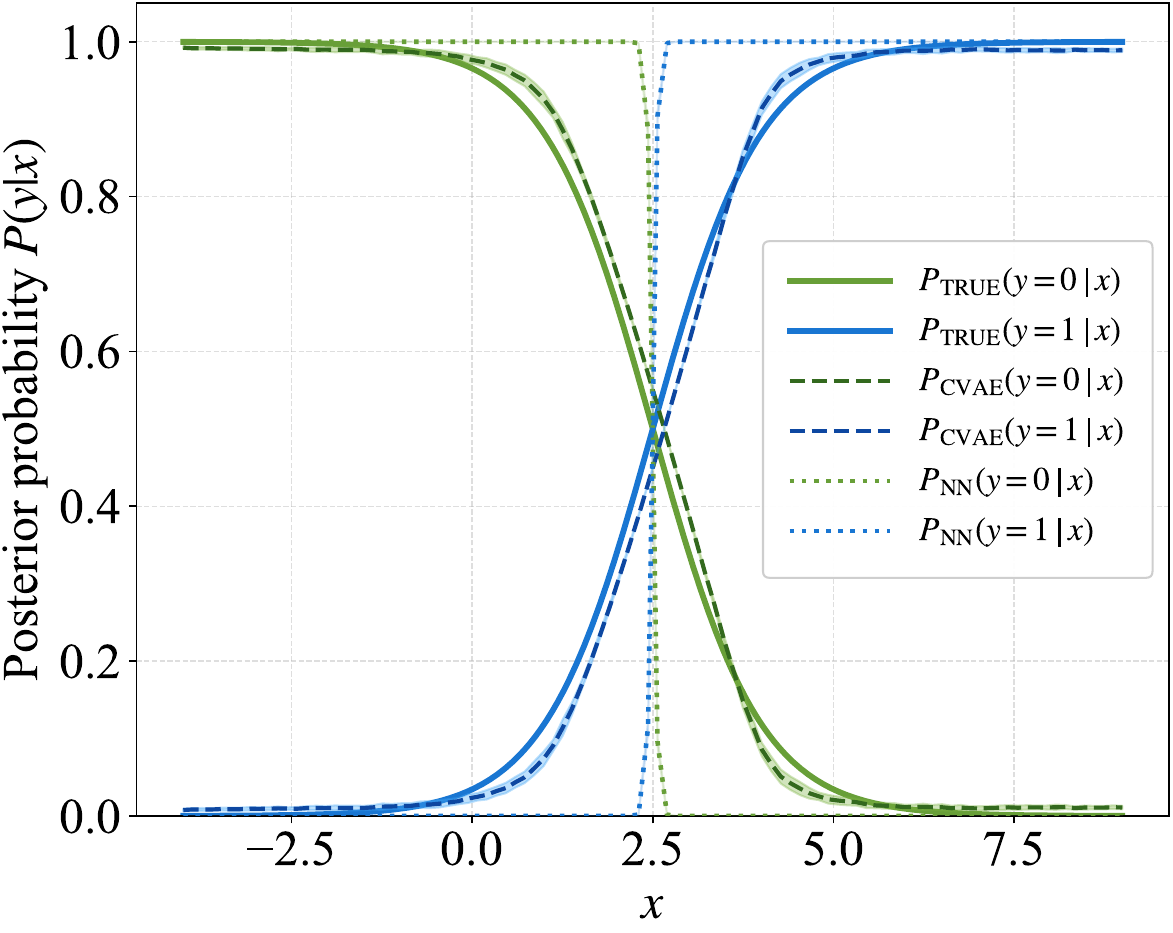}
}
\caption{\label{compare_cvae_cnn}Posterior estimate of discriminative classifier and proposed CVAE-classifier. 
}
\end{figure}

We implement the CVAE-inference procedure discussed above and obtain $P_{\text{CVAE}}(y\mid x)$. As a comparison, we also train a standard neural network classifier by minimizing the cross-entropy loss on the training data and use the predictive probability as the neural network estimate $P_{\text{NN}}(y\mid x)$. Both probabilities are plotted in Figure \ref{compare_cvae_cnn}.  The solid lines represent the true posterior probabilities $P_{\mathrm{TRUE}}$, the dashed lines represent the posterior probabilities estimated by the CVAE-based classifier, denoted by $P_{\mathrm{CVAE}}$, and the dotted lines represent the posterior probabilities $P_{\mathrm{NN}}$ predicted by the discriminative classifier. 

The posterior probability of those classifiers exhibits different trends. 
Although the discriminative model can accurately identify the decision boundary, the posterior probability deviates from the true posterior distribution when $x$ is outside the training set, suggesting that the NN-classification model is not well-calibrated in out-of-distribution inputs \citep{guo2017calibration,lakshminarayanan2017simple}. Such findings echo with the literature that indicates prediction distributions generated by powerful NNs tend to be overconfident \citep{nguyen2015deep}.

In contrast, the posterior estimate of the proposed CVAE-classifier is close to the ground truth posterior distribution, indicating the proposed generative classifier can calibrate posterior probability on out-of-distribution samples better.

\subsection{The Prior Model}\label{sec-prior}
This section will introduce the modeling of $p(\bz_t\mid \bz_{t-1})$ in the proposed GBF framework. Recall that the complete latent variable is defined as $\bz_t=[\bc_t,\beps_t]$, where the condition label $\bc_t$ and latent continuous embedding $\beps_t$ capture class-specific patterns and within-class variations in the high-dimensional signals, respectively.
It is natural to assume that their prior distributions are independent $p(\bz_t\mid \bz_{t-1})=p(\bc_t\mid \bz_{t-1})p(\beps_t\mid \bz_{t-1})$. Hence, we construct models separately for the latent components: $p(\beps_t \mid \bc_{t-1}, \beps_{t-1})$ and $p(\bc_t \mid \bc_{t-1}, \beps_{t-1})$. 

\textbf{Continuous embedding modeling}: For the continuous embedding, we simply assume it is independent over time and follows a standard Gaussian distribution, as is standard in CVAE formulations: 
\begin{equation}
\label{eqn:continuous_embedding}
\beps_t \mid \bc_{t-1}, \beps_{t-1} \sim \mathcal{N}(0, I).
\end{equation}
This assumption is consistent with the role of $\beps_t$ as an auxiliary latent variable that captures within-class variations in the observation space. Under the Gaussian prior, the negative log-prior of $\beps_t$ is proportional to $\| \beps_t\|^2$ up to a constant.

\textbf{Class variable modeling}: For the discrete class variable, there are multiple possible ways to incorporate prior information. 
In this paper, since the system state is represented as a finite set of discrete classes, we model the class dynamics using a Markov transition model.

Let $\mathbf{s}_t \in \{e_1, \dots, e_K\} \subset \mathbb{R}^K$ denote the true discrete state of the system at time $t$, where $e_i$ is the $i$-th canonical basis vector. The system state is represented by a probability vector $\bc_t$, which describes the distribution of $\mathbf{s}_t$ such that $[\bc_t]_i = \mathbb{P}(\mathbf{s}_t = e_i)$, with $[\bc_t]_i \ge 0$ and $\sum_{i=1}^K [\bc_t]_i = 1$. 
The state dynamics could be represented by a transition matrix $A \in \mathbb{R}^{K \times K}$, where:
\begin{equation}
    A_{ij} = \mathbb{P}(\mathbf{s}_t = e_j \mid \mathbf{s}_{t-1} = e_i), \quad A_{ij} \ge 0, \quad \sum_{j=1}^K A_{ij} = 1.
\end{equation}
The transition probability matrix $A$ could be estimated from historical system trajectories and serves as the temporal prior for filtering.
Given the previous belief state $\bc_{t-1}$, the prior belief over the state at time $t$ is computed as 
\begin{equation}
\label{prediction}
\tilde{\bc}_t = A^\top \bc_{t-1}.
\end{equation}
This predicted belief parameterizes the categorical prior distribution of the current discrete state:
\begin{equation}
\bc_t \mid \bc_{t-1}, \beps_{t-1} \sim \text{Categorical}(\tilde{\bc}_t),
\end{equation}
where $\tilde{\bc}_t = \mathcal{T}(\bc_{t-1})$ represents the prior belief dynamically predicted from the previous state. 
Therefore, the corresponding categorical log-prior can be written as
\begin{equation}\label{eqn:logprior}
    \log p(\bc_t \mid \bc_{t-1}, \beps_{t-1}) = \sum_{k=1}^K [\bc_t]_k \log [\tilde{\bc}_t]_k,
\end{equation}
where $K$ is the total number of classes, and $[\cdot]_{k}$ denotes the $k$-th components of the posterior state probability vector. For notational simplicity, we define the soft cross-entropy loss between two probability distributions $p$ and $q$ as $\mathcal{L}_{\text{softCE}}(p, q) := -\sum_{i=1}^K p_i \log q_i$, which is the negative of the log prior defined in   \eqref{eqn:logprior}.

\subsection{Score-based Posterior Sampling}

With the prior model and the likelihood model, sampling from the posterior $p(\bz_t \mid \bx_t, \bz_{t-1})$ still appears challenging because the partition function $p(\bx_t \mid \bz_{t-1})$ is often difficult to compute in practice.
To circumvent the difficulty of calculating the partition function, we take advantage of the fact that the partition function is independent of the current latent state $\bz_t$ that we want to infer. Therefore, a natural solution is to use score-based sampling approaches. The score function is defined as the gradient of the log-posterior:
\begin{equation}
\label{eqn:logposterior}
\begin{aligned}
   \nabla_{\bz_t} \log p(\bz_t \mid \bx_t, \bz_{t-1})= \nabla_{\bz_t} \log p(\bx_t \mid \bz_t) + \nabla_{\bz_t} \log p(\bz_t \mid \bz_{t-1}). 
\end{aligned}
\end{equation}
It comprises two parts: the gradient of the observation model and the gradient of the prior model. It is worth noting that the intractable partition function does not appear in the score function as it is not dependent on $\bz_t$. Hence, the score of $p(\bz_t \mid \bx_t, \bz_{t-1})$ are tractable through backpropagation. 

Based on the probabilistic formulation of the CVAE model in section \ref{sec-cave}, 
the conditional density is thus given by $
    p_{\theta}(\bx_t \mid \bz_t) \propto \exp\left(-\frac{1}{2\sigma^2} \|\bx_t - g_{\theta}(\bc_t, \beps_t)\|^2 \right)$. Accordingly, the negative log-likelihood is $
    -\log p_{\theta}(\bx_t \mid \bc_t, \beps_t) = \frac{1}{2\sigma^2} \|\bx_t - g_{\theta}(\bc_t, \beps_t)\|^2 + \text{const}$, 
which corresponds to a $\ell_2$ reconstruction objective. 
Combined with the prior model in section \ref{sec-prior}, the gradients of the log-posterior with respect to the latent variables become:
\begin{equation}
\label{eqn:scorec}
\begin{aligned}
    \nabla_{\bc_t} \log p(\bz_t\mid \bx_t,\bz_{t-1}) 
    \propto \nabla_{\bc_t} \left( -\frac{1}{2\sigma^2} \|\bx_t - g_{\theta}(\bc_t, \beps_t)\|^2 - \mathcal{L}_{\text{softCE}}(\bc_t,\tilde{\bc}_t) \right),
\end{aligned}
\end{equation}
and
\begin{equation}
\label{eqn:scoreeps}
\begin{aligned}
 \nabla_{\beps_t} \log p(\bz_t\mid \bx_t,\bz_{t-1}) 
 \propto \nabla_{\beps_t} \left( -\frac{1}{2\sigma^2} \|\bx_t - g_{\theta}(\bc_t, \beps_t)\|^2-\frac{1}{2} \|\beps_t\|^2 \right).   
\end{aligned}
\end{equation}
To further simplify notation, we can define a total energy function $\mathcal{L}_{\mathrm{total}}$ as:
\begin{equation}
\label{eqn:totallossdef}
       \mathcal{L}_{\mathrm{total}}(\bc_t,\beps_t)
    :=
    \left\|
    \bx_t - g_{\theta}(\bc_t,\beps_t)
    \right\|^2
    +\mu_{\text{latent}} \cdot \sigma^2 \cdot \|\beps_t\|^2
    +
    2\mu_{\text{prior}} \cdot \sigma^2  \cdot \mathcal{L}_{\mathrm{softCE}}(\bc_t,\tilde{\bc}_t).
\end{equation}

where $\mu_{\text{latent}}$ and $\mu_{\text{prior}}$ are weighting factors that balance the observation likelihood against the temporal prior. It is clear that the total loss $\mathcal{L}_{\text{total}}$ defined in ~\eqref{eqn:totallossdef} is proportional to the negative log posterior defined in ~\eqref{eqn:logposterior} , that is, $\mathcal{L}_{\text{total}}\propto -\log p(\bz_t\mid \bx_t,\bz_{t-1})$.

\textbf{Stochastic Gradient Langevin Dynamics}: Since the gradient of $\mathcal{L}_{\text{total}}$ can be efficiently calculated via auto-differentiation, we leverage Stochastic Gradient Langevin Dynamics (SGLD)~\citep{welling2011bayesian} to generate samples from $\exp(-\mathcal{L}_{\text{total}})$.  More specifically, we parametrize $\bc_t$ by logits $\bxi_t$ such that $\bc_t=\mathrm{softmax}(\bxi_t)$, and apply the following gradient-based update rule,
\begin{equation}
\label{eqn:sgldupdate}
\left\{\begin{aligned}
    \bxi_t &\leftarrow \bxi_t - \eta \nabla_{\bxi_t}\mathcal{L}_{\text{total}}+\sqrt{\eta}\varepsilon_{\bxi_t}\\
    \beps_t &\leftarrow \beps_t - \eta \nabla_{\beps_t}\mathcal{L}_{\text{total}}+\sqrt{\eta}\varepsilon_{\beps_t},\\
\end{aligned}\right.
\end{equation}
where $\eta$ is the learning rate, gradients are proportional to the score functions~\eqref{eqn:scorec} and~\eqref{eqn:scoreeps} and calculated by auto-differentiation, and $\varepsilon_{\beps_t}\in\mathbb{R}^{d_{\beps}}$ and $\varepsilon_{\bxi_t}\in\mathbb{R}^{d_{\bxi}}$ are sampled from multivariate standard normal distributions. The convergence of SGLD has been discussed in literature~\citep{sgldconverge,kinoshita2022improved}. 

In practice, we substitute the standard gradient descent step in SGLD with its adaptive step-size variant~\citep{kim2020stochastic}, AdamWSGLD, to ensure faster numerical convergence while maintaining the exploratory benefits of the injected noise. Similar to particle filtering, we use $N$ trajectories starting from $N$ randomly initialized particles instead of a single trajectory of SGLD to improve the coverage of the trajectories~\citep{lakshminarayanan2017simple}.

\subsection{Recursive Bayesian Filtering}

The score-based posterior sampling procedure completes the state estimation for a single time step. To apply the proposed method to streaming observations, we embed it into a recursive Bayesian filtering pipeline, where the posterior belief obtained at the current time step is propagated forward as the prior belief for the next time step.
The pseudocode for Generative Bayesian Filtering is summarized in Algorithm \ref{alg:gbf}. 

\begin{algorithm}[!h]
\caption{Generative Bayesian Filtering}
\label{alg:gbf}
\begin{algorithmic}[1]
\REQUIRE Previous state belief $\bc_{t-1}$, new observation $\bx_t$, transition matrix $A$, pretrained encoder $q_{\phi}$ and decoder $g_{\theta}$, weight factor $\mu_{\text{latent}},\mu_{\text{prior}}$, number of candidate samples $N$
\ENSURE Posterior state estimate $\bc_t$

\STATE \textbf{Prediction step}: Compute the prior belief via the transition model: $\tilde{\bc}_t = A^\top \bc_{t-1}$ 

\STATE \textbf{Correction Step}: Initialize $N$ latent variable pairs $(\bxi^{(i)},\beps^{(i)})\sim (0,  q_{\phi}(\beps_t \mid \bx_t, \bc^{(i)}))$.
\FOR{step $m=1$ to $M$}
    \FOR{ sample $i=1$ to $N$}
        \STATE Variable transform:
        $\bc^{(i)} = \mathrm{softmax}(\bxi^{(i)}) $
        \STATE Reconstruct the observation from the latent variables:
        $ \hat{\bx}^{(i)} = g_\theta(\bc^{(i)}, \beps^{(i)}) $
    \STATE Compute the reconstruction loss: $\mathcal{L}_{\text{recon}} =  \|\bx_t - \hat{\bx}^{(i)}\|^2$
    \STATE Compute the prior consistency loss:
    $ \mathcal{L}_{\text{prior}} = 
     \mu_{\text{latent}} \cdot \|\beps_t^{(i)}\|^2+ \mu_{\text{prior}} \cdot \mathcal{L}_{\text{softCE}}(\bc^{(i)}, \tilde{\bc}_t)$
    \STATE Compute total loss: $\mathcal{L}_{\text{total}} = \mathcal{L}_{\text{recon}} + \mathcal{L}_{\text{prior}}$
    \STATE Update $(\bxi^{(i)}, \beps^{(i)})$ via SGLD based on the gradient 
    $\nabla\mathcal{L}_{\text{total}}$ by~\eqref{eqn:sgldupdate}.
\ENDFOR
\ENDFOR
\STATE Average the optimized state probabilities:
$ \bc_t = \frac{1}{N} \sum_{i=1}^N \bc^{(i)} $
\RETURN $\bc_t$
\end{algorithmic}
\end{algorithm}

Each Bayesian filtering step consists of a prediction step from prior belief and a correction step of $\bc_t$ from score-based posterior sampling. Upon completion of the posterior sampling loop, we aggregate the sampled particles by calculating their empirical mean, $\bc_t = \frac{1}{N} \sum_{i=1}^N \bc^{(i)}$, to establish the final posterior estimate of the latent state at time $t$. This framework is thus inherently recursive. The newly computed posterior $\bc_t$ serves directly as the prior belief for the subsequent time step $t+1$ by replacing $\bc_{t-1}$ in   \eqref{prediction}. As new observations $\bx_{t+1}, \bx_{t+2}, \dots$ stream in online, the algorithm continuously cycles through the dynamics-driven prediction stage and the generation-driven correction stage to enable robust and real-time tracking of the system's underlying states.

\section{Numerical Experiments}\label{sec-num}

In this section, we construct a temporal MNIST dataset and a temporal Fashion MNIST dataset where images evolve with synthetic stochastic dynamics and evaluate the numerical performance of Algorithm~\ref{alg:gbf}. 
To comprehensively assess the performance of the GBF framework, we compare the proposed GBF against several representative baseline methods, including:
\begin{itemize}
    \item \textbf{CVAE} (proposed): the CVAE inference scheme introduced in section~\ref{sec-cave}. This implementation does not consider the temporal information.
    \item \textbf{NN}~\citep{lecun1998gradient}:
    the standard convolutional neural network for classifying the latent state based on the current observation.
    \item \textbf{HMMNN}~\citep{yang2021single}: a discriminative algorithm that combines a neural network classifier with a Hidden Markov Model, treating neural-network predictions as observation sequences and leveraging transition prior for state inference. This baseline is included to evaluate the benefit of adding temporal dynamics to a discriminative classifier.
    \item \textbf{SCALE}~\citep{oraki2026scale}: a heuristic approach that uses a conditional VAE to model class-conditional feature distributions, then ranks candidate classes by ELBO-derived energy function for classification. This method allows us to compare GBF with representative generative inference models.
    \item \textbf{DEEPBAYES}~\citep{li2019generative}: a technique that uses a deep latent variable model as a generative classifier. DEEPBAYES marginalizes $\beps$ rather than optimizing it. Thus the comparison between GBF and DEEPBAYES demonstrates the effectiveness of the proposed SGLD sampler.
    \item \textbf{KALMAN}~\citep{kalman1960new}: the standard Kalman filtering method in which class label vectors are treated as latent dynamic states. The classical filtering method is chosen to assess the limitations of linear-Gaussian assumptions in state estimation.
\end{itemize}

MNIST \citep{lecun1989backpropagation} is a widely used benchmark consisting of handwritten digit images from ten classes, with each image represented by a $28 \times 28$ pixel array. To evaluate the proposed framework, we construct a synthetic temporal classification task based on the MNIST dataset. Specifically, each digit category is treated as a discrete system state, and temporal MNIST sequences are artificially generated according to a predetermined Markov model. Here, we only adopt digits $1\sim4$ as our states. 
Once the state is determined, an MNIST image belonging to the corresponding digit class is randomly sampled as the observation at that time step. In this way, the original static image dataset is converted into a sequential dataset with temporal dependence.
We similarly build a temporal Fashion-MNIST dataset by choosing \textit{T-shirt, Trouser, Dress, Sneaker, Bag} in the Fashion-MNIST dataset \citep{xiao2017fashion} as the state of each step, and sampling images with respect to the state.

\begin{figure}[htbp!]
    \centering
    \includegraphics[width=0.8\linewidth]{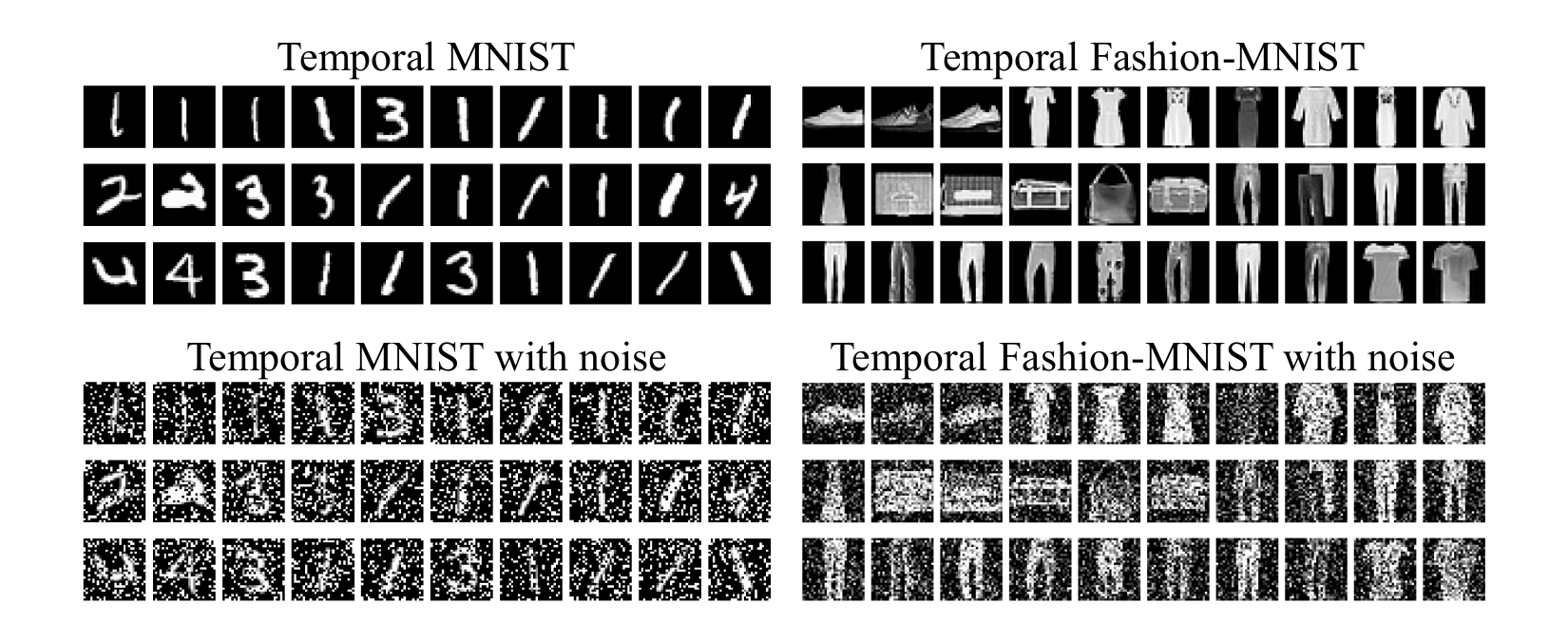}
    \caption{Temporal MNIST and temporal Fashion-MNIST example.}
    \label{fig:mnist}
\end{figure}

In practical engineering systems, the signals from sensors are often contaminated by noise and corruption that arise from imperfect acquisition conditions, environmental disturbances, and hardware limitations. Such observation degradation may significantly reduce the reliability of state estimation if not properly accounted for.
To further mimic the uncertainty and noise in practical sensing scenarios, we corrupt the temporal MNIST dataset by random pixel flipping. Specifically, for each image, every pixel is binarized and independently flipped with probability $p_{\mathrm{flip}} \in [0,1]$, which is utilized to quantify the level of observation noise. For the temporal Fashion MNIST dataset, Gaussian noise is added to the original images.
An example of the constructed temporal sequence and noisy version is shown in Figure \ref{fig:mnist}.

In order to implement the GBF, we first train a CVAE model offline, which serves as the generative observation model in the filtering process. In the training process, the condition variable $\bc$ is represented by one-hot coding. The encoder and decoder are both built with a CNN-based backbone to capture the spatial structure of images in the dataset. The length of each time series MNIST data is 500, and the number of sequence for training and testing is 50 and 8. Similar to the temporal MNIST dataset, the length of the Fashion-MNIST sequence is 500, and 60 sequences are used for training, while 10 sequences are left for testing. 
After offline training, the decoder is fixed and incorporated into the GBF inference procedure as the generative observation model. 

\begin{figure}[h]
    \centering
    \includegraphics[width=0.5\textwidth]{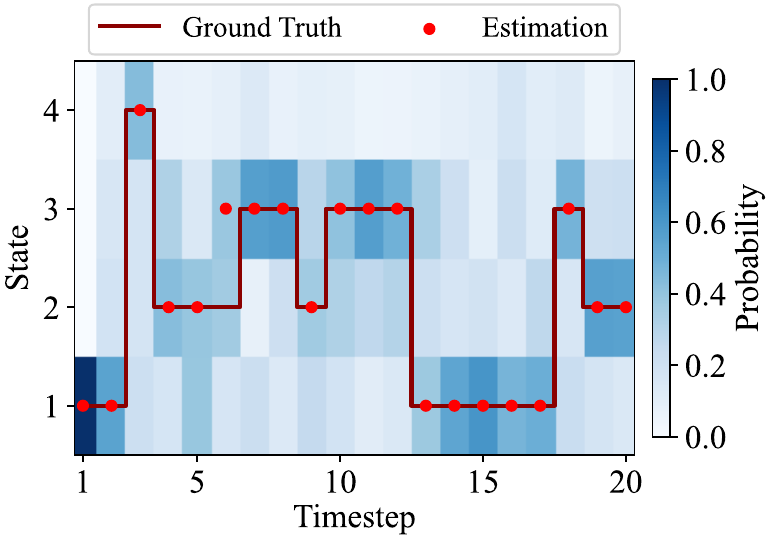}
    \caption{Posterior state estimation of GBF on temporal MNIST sequence.}
    \label{fig:heatmap}
\end{figure}

\noindent\textbf{Posterior state estimate} 
Figure \ref{fig:heatmap} visualizes how the posterior state probabilities evolve over time with the temporal MNIST dataset under a 0.15 noise level. The red step curve denotes the ground-truth latent state, and the red dots denote the state estimated at each time step. Overall, the estimated states closely follow the ground truth, and the posterior probability mass is generally concentrated on the correct state, showing the GBF could track the temporal evolution of the latent state. Moreover, a more dispersed posterior distribution indicates increased uncertainty, which is often associated with misclassification of the latent state. Therefore, the proposed method not only provides point estimates of the latent state but also produces informative posterior uncertainty for estimation.

\begin{figure}[h]
    \centering
    \includegraphics[width=0.8\linewidth]{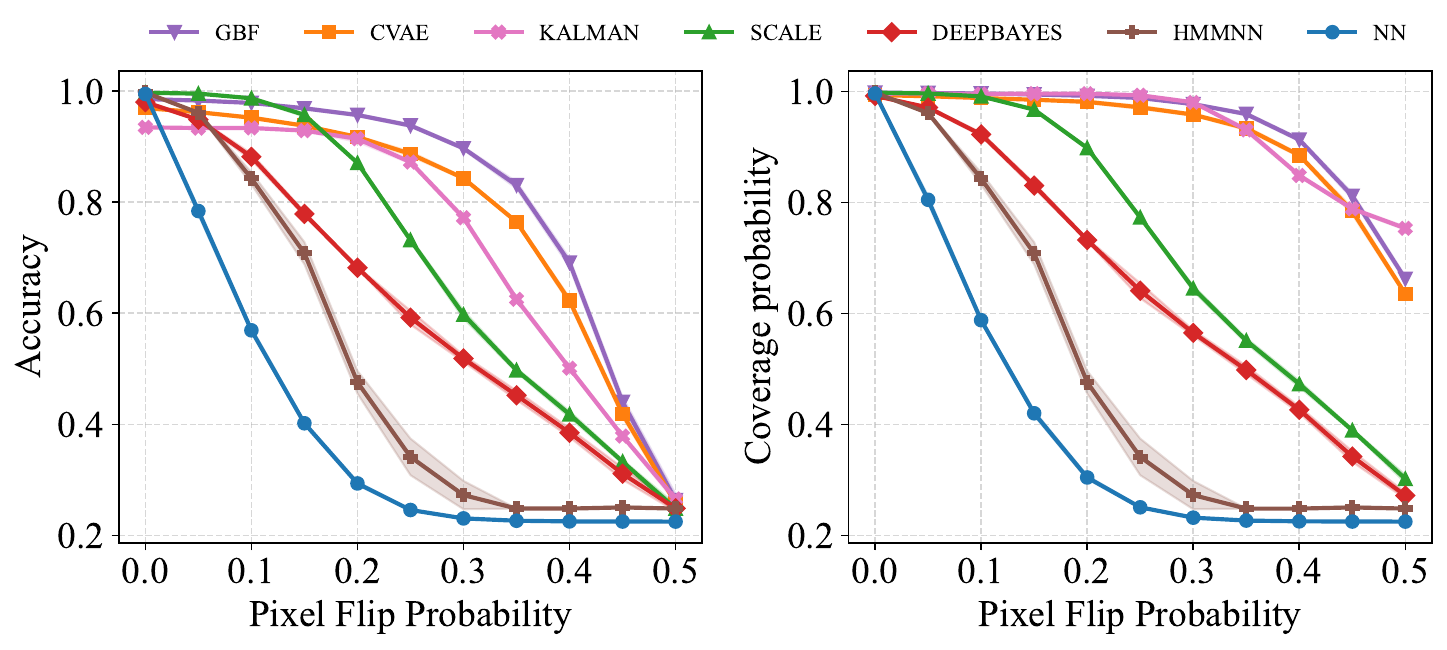}
    \caption{Accuracy and coverage probability of temporal MNIST dataset under different pixel flip probabilities.}
    \label{fig:mnist_result}
\end{figure}

\begin{figure}[h]
    \centering
    \includegraphics[width=0.8\linewidth]{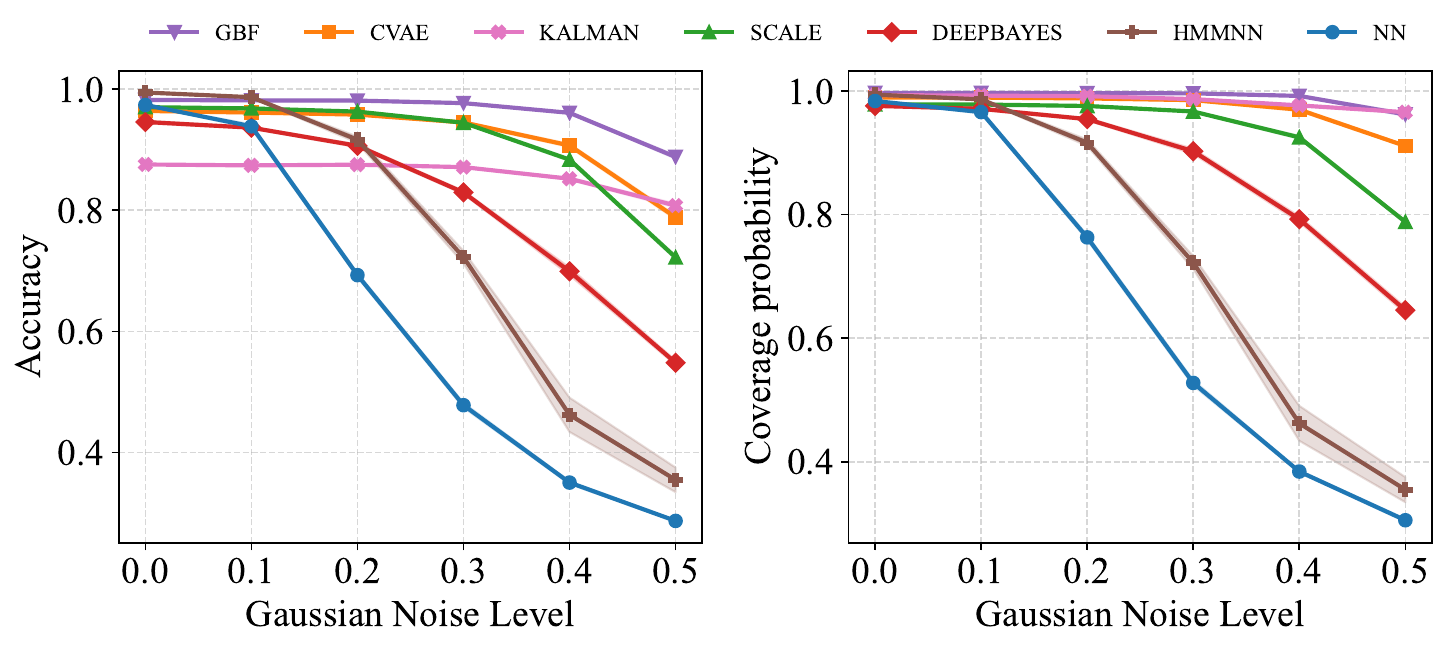}
    \caption{Accuracy and coverage probability of temporal Fashion-MNIST dataset under different noise levels. }
    \label{fig:fmnist_result}
\end{figure}

\noindent\textbf{Accuracy and coverage} We further evaluate the accuracy and coverage probability 
of all methods on the test dataset, and the results of temporal MNIST and temporal Fashion-MNIST are reported in Figure \ref{fig:mnist_result} and Figure \ref{fig:fmnist_result}, respectively. 
Here we adopt coverage probability to evaluate the uncertainty quantification performance. Let $\hat{\mathbf{c}}_i\in\mathbb{R}^K$ denote the predicted class-probability vector over $K$ latent states at time step $i$. First sort the entries of $\hat{\mathbf{c}}_i$ from the largest to the smallest, and let $\pi_i$ be the corresponding permutation such that $\hat c_{i,\pi_i(1)}
\ge
\hat c_{i,\pi_i(2)}
\ge
\cdots
\ge
\hat c_{i,\pi_i(K)} .$ Given a threshold $\tau$, $k_i(\tau)$ is defined as the smallest integer such that the cumulative predicted probability of the top $k_i(\tau)$ classes reaches $\tau$:
\begin{equation}
k_i(\tau)
=
\min
\left\{
m:
\sum_{r=1}^{m}
\hat c_{i,\pi_i(r)}
\ge \tau
\right\}.
\end{equation}
The corresponding prediction set is
$
\mathcal{C}_i(\tau)
=
\left\{
\pi_i(1),\ldots,\pi_i(k_i(\tau))
\right\}$, and the coverage probability could be calculated as
\begin{equation}
\mathrm{Coverage}(\tau)
=
\frac{1}{N}
\sum_{i=1}^{N}
\mathbf{1}
\left[
y_i \in \mathcal{C}_i(\tau)
\right],
\end{equation}
where $y_i$ is the ground-truth class label, and $\tau=0.75$ is chosen in this paper.
It is worth noting that all the methods are trained on clean data without any noise, and then tested on the test dataset with varying noise levels. All methods achieve high accuracy and coverage probability with clean data. However, the performance of the discriminative method degrades significantly as the noise increases, which shows the sensitivity of the NN-based classifier to input noise. This is not surprising as NNs trained on clean data are known to be brittle to distribution shift~\citep{goodfellow2014explaining}. By incorporating temporal dynamics through a hidden Markov model, the HMMNN method consistently maintains better performance than NN. Besides, generative classifiers (GBF, CVAE, SCALE, DEEPBAYES) remain stable under even moderately high probabilities of pixel flip, demonstrating the robustness of the generative observation model. 
Among the generative methods, GBF achieves the best overall performance by combining the flexible observation model with prior information. Compared with SCALE, which ranks candidate classes using an ELBO-derived energy score, GBF performs a Bayesian posterior update and provides a recursive filtering formulation for temporal state estimation. DEEPBAYES estimates system states through marginalization over the continuous latent variable, while our proposed method utilizes SGLD to jointly sample and refine the categorical state variable and the continuous latent variable, allowing the posterior estimate to retain within-class variability and uncertainty information. This sampling-based posterior refinement leads to more robust state estimation under noisy observations.


To further evaluate the quality and reliability of probabilistic predictions, we also calculate the Brier score and expected calibration error (ECE). The definition of metrics and results of the two datasets are summarized in the \nameref{appendix}. The Brier score and ECE results show that the generative methods provide more reliable probabilistic predictions under corrupted observations, and the traditional methods tend to be overconfident with clean data.

\section{ Case Studies}\label{sec-case}
We further evaluate the performance of the proposed GBF on two real world sensing problems, including additive manufacturing process monitoring and ECG-based cardiac state diagnosis, both of which are characterized by high-dimensional and noisy observation signals.

\subsection{Case Study I: Additive Manufacturing State Monitoring}\label{lpbf}
Laser powder bed fusion (LPBF) is a fundamental metal processing method in additive manufacturing \citep{ren2025mechanistic}, where a laser selectively melts thin layers of metal powder to form dense three-dimensional parts. LPBF has been widely used in advanced manufacturing and is famous for its capability to fabricate refractory metals into complex components.

\begin{figure}[htbp!]
    \centering
    \includegraphics[width=0.45\linewidth]{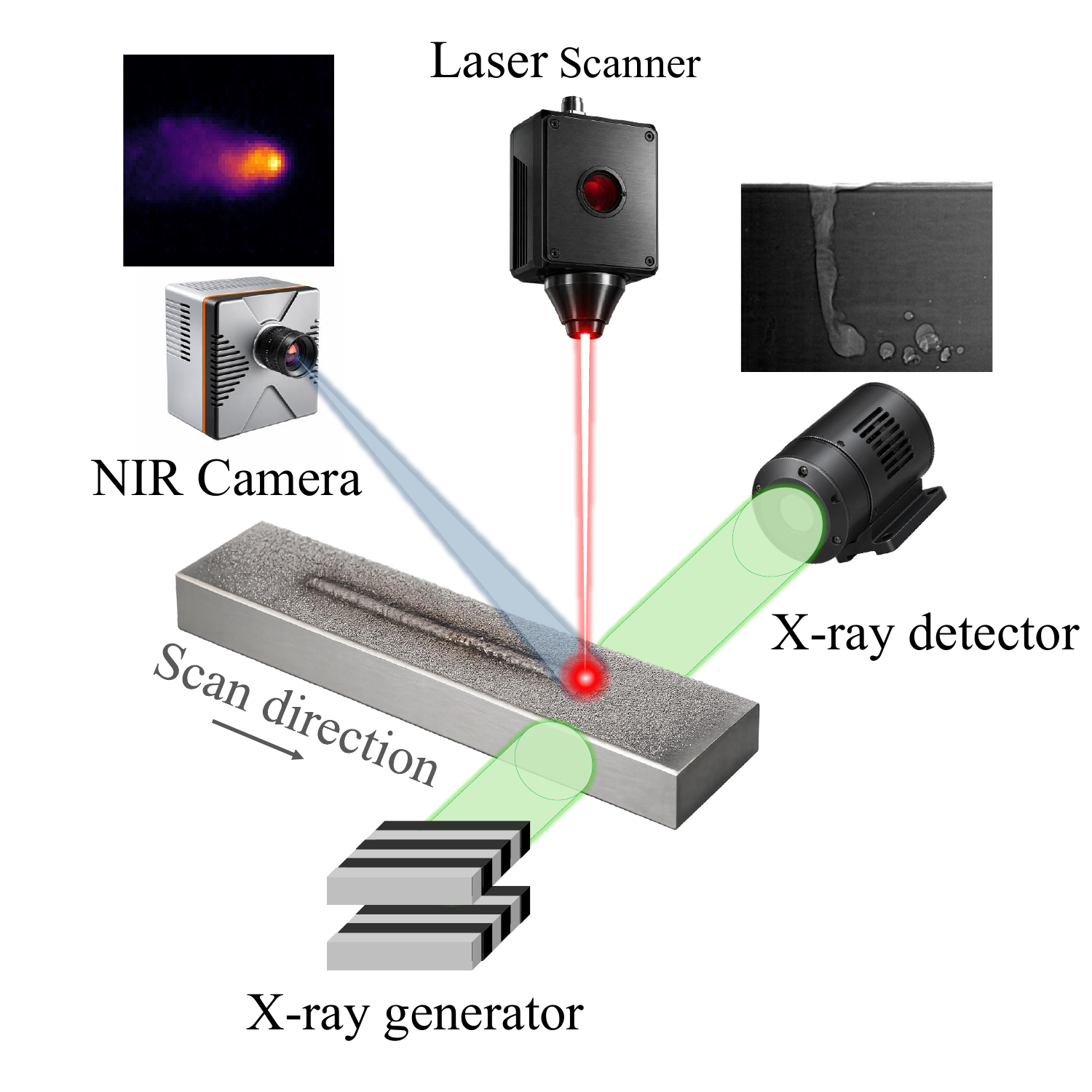}
    \caption{Experimental setup of laser powder bed fusion.}
    \label{fig:lpbf_setup}
\end{figure}

However, during the LPBF process, multiple defects can arise~\citep{yin2025artificial}. An important defect is the pore in the metal part, which can concentrate stress and degrade the mechanical properties of the build. Previous studies have shown that these pores originate from unstable keyhole dynamics, where gas bubbles detach from the recoil pressure and then are trapped during rapid solidification \citep{ren2023machine}. To accurately identify pore-generation events, in-situ synchrotron imaging is an ideal tool because it can directly visualize the subsurface dynamics and provide high spatial and temporal resolution for model development and validation \citep{cunningham2019keyhole}. However, the high cost and complexity of synchrotron imaging equipment make the industrial deployment unrealistic. Compared to synchrotron imaging, the near-infrared (NIR) sensing technique is deployable on the manufacturing shop floor and captures the thermal radiation emitted from the surface of the melt pool and the keyhole region. Hence, NIR sensors provide in-situ monitoring signals for LPBF. The schematic diagram of our experimental setup is shown in Figure \ref{fig:lpbf_setup}, where the synchrotron X-ray images are used to obtain the system state as the ground truth label to train the CVAE, and NIR images serve as partial observation for state monitoring.

\begin{figure}[htbp!]
    \centering
    \includegraphics[width=0.7\linewidth]{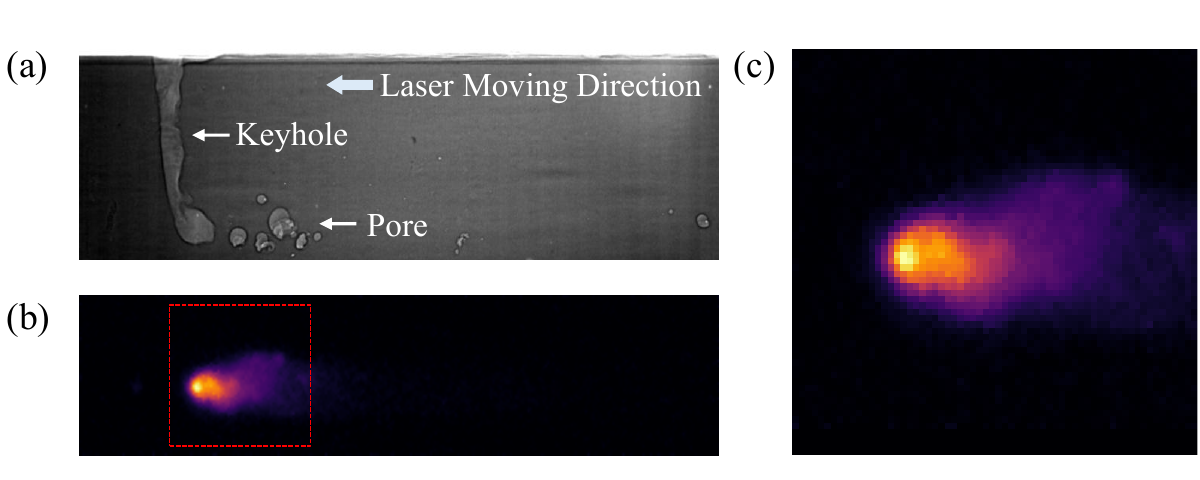}
    \caption{Measurement and preprocessing of LPBF.}
    \label{fig:lpbf_data}
\end{figure}

We collected 74 trajectories of LPBF processing with synchrotron X-ray and near-infrared (NIR) measurements \citep{ren2024sub}. The X-ray image, as shown in Figure \ref{fig:lpbf_data}(a), is acquired at 50 kHz, while the NIR data shown in Figure \ref{fig:lpbf_data}(b) is recorded at 250 kHz. Using the X-ray observations as the ground-truth reference, each frame is labeled as either pore or non-pore according to whether a pore generation is observed. To further capture the dynamic nature of the process, the NIR data were segmented using time windows, and each window was matched with the corresponding X-ray-based label. In addition, preprocessing was applied to the NIR frames. As described in Figure \ref{fig:lpbf_data}(c), for each time step, the region of interest is cropped around the melt pool, with the crop center shifted along the laser moving direction to retain trailing-region information.

Similarly, to implement the proposed framework on the LPBF state monitoring task, we first train a CVAE model based on the collected dataset. Among the 74 trajectories, 65 are used for model training, while the remaining trajectories are reserved for evaluation. Compared with the synthetic dataset, the LPBF monitoring problem involves more complex sensing patterns and stronger within-class variations. Therefore, the dimension of the latent variable $\beps$ is increased to 128 in order to provide a stronger representation capacity for modeling the high-dimensional NIR observations.

Since the task is to detect whether a new pore is generated in given time windows, we use accuracy, F1 score, and area under the curve (AUC) to evaluate the performance of different methods on this binary classification task. The results are listed in Table \ref{tab:classifier_comparison}. 

\begin{table}[ht]
\centering
\caption{Classification performance on the LPBF dataset.}
\label{tab:classifier_comparison}
\renewcommand{\arraystretch}{1.15}
\setlength{\tabcolsep}{14pt}
\arrayrulecolor{black}
\footnotesize
\resizebox{0.6\textwidth}{!}{

\begin{tabular}{lccc}
\toprule
\textbf{Methods} 
& \textbf{Accuracy $\uparrow$} 
& \textbf{F1 $\uparrow$} 
& \textbf{AUC $\uparrow$} \\
\midrule
\textbf{GBF}      
& \best{0.853 \pm 0.008} 
& \best{0.875 \pm 0.007} 
& \val{0.837}{0.016} \\

\textbf{CVAE}     
& \val{0.823}{0.005} 
& \val{0.846}{0.006} 
& \val{0.839}{0.024} \\

\textbf{KALMAN}   
& \val{0.681}{0.000} 
& \val{0.800}{0.000} 
& \best{0.855 \pm 0.000} \\

\textbf{SCALE}    
& \val{0.798}{0.000} 
& \val{0.819}{0.000} 
& \val{0.794}{0.000} \\

\textbf{DEEPBAYES}
& \val{0.617}{0.032} 
& \val{0.608}{0.038} 
& \val{0.624}{0.032} \\

\textbf{HMMNN}    
& \val{0.800}{0.012} 
& \val{0.825}{0.013} 
& \val{0.794}{0.012} \\

\textbf{NN}       
& \val{0.798}{0.000} 
& \val{0.826}{0.000} 
& \val{0.808}{0.001} \\
\bottomrule
\end{tabular}
}
\end{table}

GBF achieves the best accuracy and F1 score among all methods. Although Kalman filtering obtains the highest AUC, its accuracy is much lower, suggesting that the simple linear-Gaussian observation model is insufficient for complicated sensor data. The good performance of CVAE indicates that generative models could infer states effectively in the LPBF monitoring task by learning the nonlinear and high-dimensional NIR signals. Moreover, the improvement of GBF over CVAE suggests temporal transition information is useful for pore-generation detection, because the states of neighboring windows in LPBF are correlated. The generative methods are better than discriminative methods overall, which suggests NN is more sensitive to sensor noise and data variability. Also, heuristic generative inference baselines like SCALE and DEEPBAYES harvest the benefits of generative models, but do not fully exploit the posterior distribution, which could lead to suboptimal performance.

\subsection{Case Study II: ECG-based Cardiac State Monitoring}\label{ecg}
\begin{figure}[htbp!]
    \centering
    \includegraphics[width=0.5\linewidth]{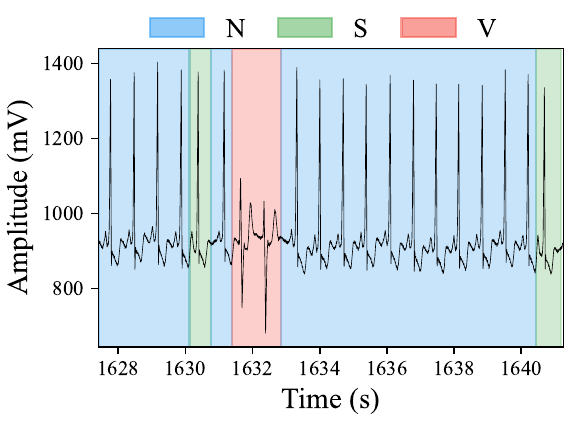}
    \caption{Example of ECG signal in MIT-BIH dataset.}
    \label{fig:ecg_sample}
\end{figure}

Electrocardiogram(ECG)-based cardiac abnormal state detection extracts information for long-term health management and clinical decision support \citep{obermeyer_ecg_2026}.
As the most common non-invasive signal to monitor the electrical activity of the heart, ECG provides a cost-effective modality to capture abnormal cardiac patterns. However, the ECG signals are susceptible to noise. Additionally, since ECG signal records accumulate over time, manual inspections of long-term ECG recordings become labor-intensive. 
Motivated by these challenges, we evaluate the proposed GBF on the MIT-BIH Arrhythmia dataset \citep{moody2001impact}, which includes 48 half-hour ECG recordings. 
Following the AAMI standard, heartbeat annotations were grouped into five classes: normal beats (N), supraventricular ectopic beats (S), ventricular ectopic beats (V), fusion beats (F), and unknown beats (Q) \citep{de2004automatic}. Thus, the goal is to infer the underlying cardiac states from the measured ECG signals.
The states are dynamic: a representative signal segment from the dataset is shown in Figure~\ref{fig:ecg_sample}, illustrating the temporal evolution of ECG morphology and the corresponding transitions among cardiac states. Such dynamic patterns arise because consecutive heartbeats are physiologically dependent, and abnormal rhythms often occur in correlated episodes.

\begin{figure}[h]
    \centering
    \includegraphics[width=0.48\linewidth]{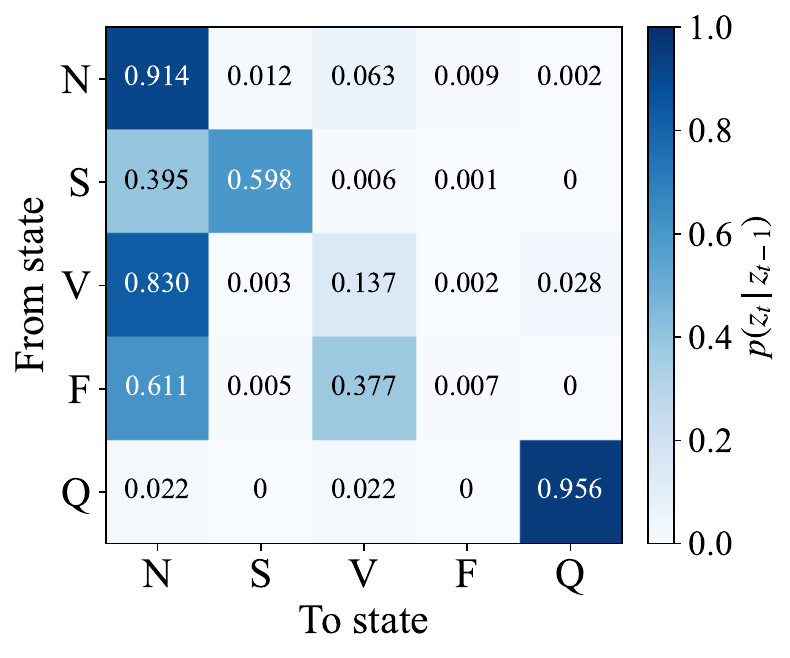}
    \caption{Transition matrix of state in the MIT-BIH dataset.}
    \label{fig:trans_matrix}
\end{figure}

 We partition the dataset at the record level, in which 38 recordings are used for training and 10 for testing. Figure~\ref{fig:trans_matrix} shows the transition matrix estimated from the dataset, in which each row represents the previous cardiac states and each column corresponds to the next state, with the entries estimated from the training set.
 
 The matrix shows that normal beats have a high probability of remaining normal, while abnormal states often transition back to the normal class. Hence, the transition matrix reveals important information of the temporal dependency of health states.

\begin{figure}[h]
    \centering
    \includegraphics[width=0.65\linewidth]{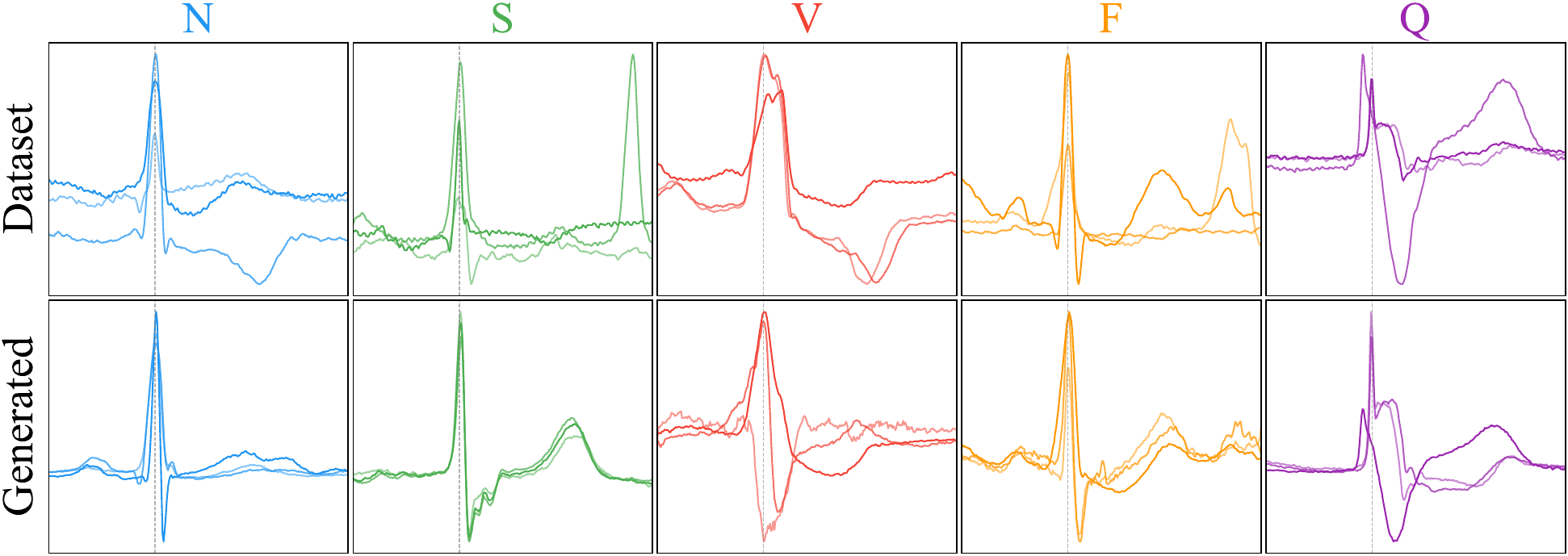}
    \caption{Original ECG signal and generated samples.}
    \label{fig:ecg_com}
\end{figure}

The original ECG signals and generated samples with various classes are plotted in Figure \ref{fig:ecg_com}. The generated samples preserve the major waveform characteristics of each class, which indicates that our generative model learns the underlying data distribution across classes. 

To simulate sensor noise, we inject additive Gaussian noise into the dataset before performing filtering. Accuracy, macro F1, and coverage probability are used to evaluate classification performance. Macro F1 is computed by first calculating the F1 score for each class and then averaging them equally across all classes, which is suitable for evaluating performance under class imbalance. The results are shown in Table \ref{tab:ecg_performance}, while the Brier score and ECE are summarized in Table \ref{tab:ecg_calibration}.

\begin{table}[h]
\centering
\caption{Classification performance comparison on the ECG dataset.}
\label{tab:ecg_performance}
\renewcommand{\arraystretch}{1.15}
\setlength{\tabcolsep}{10pt}
\arrayrulecolor{black}
\footnotesize
\resizebox{0.65\textwidth}{!}{
\begin{tabular}{lccc}
\toprule
\textbf{Methods} 
& \textbf{Accuracy $\uparrow$} 
& \textbf{Macro F1 $\uparrow$} 
& \textbf{Coverage $\uparrow$} \\
\midrule
\textbf{GBF} 
& \best{0.760 \pm 0.002} 
& \best{0.387 \pm 0.001} 
& \best{0.963 \pm 0.001} \\

\textbf{CVAE} 
& \val{0.688}{0.002} 
& \val{0.372}{0.001} 
& \val{0.899}{0.001} \\

\textbf{KALMAN} 
& \val{0.665}{0.001} 
& \val{0.357}{0.001} 
& \val{0.869}{0.000} \\

\textbf{SCALE} 
& \val{0.621}{0.002} 
& \val{0.375}{0.001} 
& \val{0.675}{0.002} \\

\textbf{DEEPBAYES} 
& \val{0.618}{0.002} 
& \val{0.250}{0.002} 
& \val{0.700}{0.001} \\

\textbf{HMMNN} 
& \val{0.618}{0.038} 
& \val{0.245}{0.014} 
& \val{0.618}{0.038} \\

\textbf{NN} 
& \val{0.611}{0.002} 
& \val{0.162}{0.002} 
& \val{0.639}{0.002} \\
\bottomrule
\end{tabular}
}
\end{table}

The results show that GBF achieves the best performance, confirming its advantages in both classification accuracy and uncertainty quantification. Since ECG signals contain nonlinear morphological variations and class-dependent waveform patterns, generative models can better capture these distributional differences among cardiac states than discriminative methods. Among generative inference methods, the proposed GBF method benefits from posterior sampling over both the categorical state and the continuous latent seed. This refinement allows the model to find a latent explanation that better reconstructs the current observation and captures within-class variability, which further improves the robustness of the method.

\section{Conclusion and Future Work}\label{sec-conc}

This paper proposes a Generative Bayesian Filtering framework for online state estimation in systems with high-dimensional signals. Different from conventional filtering methods that rely on oversimplified assumptions and modeling, the proposed approach employs a CVAE-based classifier to characterize the state-related latent variable of observations and integrates it into Bayesian recursive updating for sequential inference.

 The results of the numerical experiments and case studies indicate that GBF could achieve more accurate and robust state estimation than other methods, especially under noisy measurements.
 In the future, the GBF framework could be extended to more complicated in-situ monitoring and multi-modal sensing scenarios for system automation and decision making.

\section{Disclosure statement}\label{disclosure-statement}

The authors declare that they have no conflicts of interest.



\phantomsection\label{supplementary-material}

\bigskip

\begin{center}

{\large\bf SUPPLEMENTARY MATERIAL}

\end{center}

\begin{description}
\item[Code:] The implementation of \nameref{alg:gbf} is available at
\url{https://github.com/leicao25/Generative_Bayesian_Filtering}.

\end{description}







  \bibliography{bibliography.bib}

\section{Appendix} \label{appendix}

\subsection{Evaluation metrics}

\textbf{Brier score.} The Brier score evaluates the quality of the predicted posterior state probabilities, and it is defined as mean squared error between the predicted class probability vector and the one-hot encoded true label vector.
\[
\mathrm{BrierScore}
=
\frac{1}{N}
\sum_{i=1}^{N}
\left\|
\hat{\mathbf{c}}_i
-
\mathbf{c}_i
\right\|_2^2.
\]

\textbf{Expected Calibration Error (ECE).} Expected Calibration Error measures the discrepancy between prediction confidence and empirical accuracy.
Let $\hat{\mathbf{c}}_i \in [0,1]^K$ denote the predicted probability vector
and $\mathbf{c}_i \in \{0,1\}^K$ denote the one-hot encoded true label.
The predicted class and true class are defined as
$
\hat{y}_i = \arg\max_{k} \hat{c}_{i,k}
$ and $
y_i = \arg\max_{k} c_{i,k}.
$
The prediction confidence is
$\hat{q}_i = \max_{k} \hat{c}_{i,k}$
, and the confidence interval $[0,1]$ is divided into $M$ bins:
\[
B_m
=
\left\{
i:
\hat{q}_i
\in
\left[
\frac{m-1}{M},
\frac{m}{M}
\right)
\right\},
\qquad m=1,\dots,M.
\]
For each bin $B_m$, the empirical accuracy is
$
\operatorname{acc}(B_m)
=
\frac{1}{|B_m|}
\sum_{i\in B_m}
\mathbf{1}\{\hat{y}_i = y_i\},
$
and the average confidence is
$
\operatorname{conf}(B_m)
=
\frac{1}{|B_m|}
\sum_{i\in B_m}
\hat{q}_i.
$
In this paper, the number of bins is set as $M=5$, thus the expected calibration error is defined as
\[
\mathrm{ECE}
=
\sum_{m=1}^{M}
\frac{|B_m|}{N}
\left|
\operatorname{acc}(B_m)
-
\operatorname{conf}(B_m)
\right|.
\]

\subsection{Results of experiments}
\begin{table}[htbp]
\centering
\caption{Brier score across methods under different pixel flip probabilities of temporal MNIST. }
\label{tab:brier_score}
\renewcommand{\arraystretch}{1.15}
\setlength{\tabcolsep}{5pt}
\arrayrulecolor{black}
\resizebox{0.7\textwidth}{!}{
\begin{tabular}{c c c c c c c c}
\toprule

\textbf{Flip Ratio} 
& \textbf{GBF} 
& \textbf{CVAE} 
& \textbf{KALMAN} 
& \textbf{SCALE} 
& \textbf{DEEPBAYES} 
& \textbf{HMMNN} 
& \textbf{NN} \\
\midrule
0    & \val{0.110}{0.000} & \val{0.157}{0.000} & \val{0.394}{0.000} & \best{0.005 \pm 0.000} & \val{0.030}{0.002} & \best{0.005 \pm 0.000} & \val{0.009}{0.000} \\
0.05 & \val{0.123}{0.000} & \val{0.169}{0.000} & \val{0.424}{0.000} & \best{0.008 \pm 0.000} & \val{0.079}{0.002} & \val{0.078}{0.003} & \val{0.389}{0.011} \\
0.10 & \val{0.137}{0.001} & \val{0.186}{0.001} & \val{0.457}{0.000} & \best{0.020 \pm 0.003} & \val{0.183}{0.006} & \val{0.314}{0.022} & \val{0.813}{0.007} \\
0.15 & \val{0.155}{0.001} & \val{0.209}{0.001} & \val{0.492}{0.001} & \best{0.070 \pm 0.004} & \val{0.361}{0.006} & \val{0.584}{0.040} & \val{1.148}{0.006} \\
0.20 & \best{0.178 \pm 0.001} & \val{0.239}{0.002} & \val{0.528}{0.001} & \val{0.216}{0.004} & \val{0.541}{0.003} & \val{1.048}{0.041} & \val{1.380}{0.007} \\
0.25 & \best{0.213 \pm 0.001} & \val{0.280}{0.003} & \val{0.566}{0.001} & \val{0.461}{0.004} & \val{0.715}{0.023} & \val{1.317}{0.066} & \val{1.490}{0.002} \\
0.30 & \best{0.271 \pm 0.002} & \val{0.337}{0.001} & \val{0.605}{0.001} & \val{0.704}{0.009} & \val{0.861}{0.010} & \val{1.455}{0.051} & \val{1.532}{0.002} \\
0.35 & \best{0.357 \pm 0.003} & \val{0.425}{0.001} & \val{0.644}{0.001} & \val{0.878}{0.006} & \val{0.988}{0.014} & \val{1.503}{0.000} & \val{1.544}{0.001} \\
0.40 & \best{0.497 \pm 0.005} & \val{0.549}{0.004} & \val{0.684}{0.002} & \val{1.021}{0.010} & \val{1.124}{0.014} & \val{1.503}{0.000} & \val{1.548}{0.000} \\
0.45 & \best{0.676 \pm 0.005} & \val{0.697}{0.002} & \val{0.724}{0.001} & \val{1.174}{0.004} & \val{1.282}{0.020} & \val{1.499}{0.005} & \val{1.549}{0.000} \\
0.50 & \val{0.823}{0.005} & \val{0.845}{0.002} & \best{0.761 \pm 0.001} & \val{1.331}{0.011} & \val{1.420}{0.011} & \val{1.503}{0.000} & \val{1.550}{0.000} \\
\bottomrule
\end{tabular}
}
\end{table}

\begin{table}[htbp]
\centering
\caption{Expected calibration error under different pixel flip probabilities of temporal MNIST. }
\label{tab:ece}
\renewcommand{\arraystretch}{1.15}
\setlength{\tabcolsep}{5pt}
\arrayrulecolor{black}
\resizebox{0.7\textwidth}{!}{
\begin{tabular}{c c c c c c c c}
\toprule

\textbf{Flip Ratio}
& \textbf{GBF}
& \textbf{CVAE}
& \textbf{KALMAN}
& \textbf{SCALE}
& \textbf{DEEPBAYES}
& \textbf{HMMNN}
& \textbf{NN} \\
\midrule
0    & \val{0.208}{0.001} & \val{0.240}{0.000} & \val{0.460}{0.000} & \best{0.002 \pm 0.000} & \val{0.009}{0.001} & \val{0.003}{0.000} & \val{0.003}{0.000} \\
0.05 & \val{0.218}{0.001} & \val{0.240}{0.001} & \val{0.483}{0.001} & \best{0.003 \pm 0.000} & \val{0.008}{0.002} & \val{0.039}{0.001} & \val{0.182}{0.006} \\
0.10 & \val{0.228}{0.001} & \val{0.244}{0.002} & \val{0.507}{0.001} & \best{0.007 \pm 0.001} & \val{0.052}{0.005} & \val{0.157}{0.011} & \val{0.396}{0.004} \\
0.15 & \val{0.235}{0.003} & \val{0.249}{0.001} & \val{0.525}{0.002} & \best{0.027 \pm 0.003} & \val{0.142}{0.004} & \val{0.292}{0.020} & \val{0.565}{0.003} \\
0.20 & \val{0.242}{0.002} & \val{0.251}{0.002} & \val{0.533}{0.006} & \best{0.090 \pm 0.001} & \val{0.236}{0.001} & \val{0.524}{0.020} & \val{0.685}{0.003} \\
0.25 & \val{0.255}{0.003} & \val{0.250}{0.005} & \val{0.512}{0.002} & \best{0.205 \pm 0.004} & \val{0.326}{0.013} & \val{0.658}{0.033} & \val{0.743}{0.001} \\
0.30 & \val{0.259}{0.002} & \best{0.248 \pm 0.001} & \val{0.430}{0.004} & \val{0.324}{0.006} & \val{0.402}{0.005} & \val{0.727}{0.026} & \val{0.765}{0.001} \\
0.35 & \val{0.257}{0.005} & \best{0.225 \pm 0.003} & \val{0.295}{0.004} & \val{0.409}{0.004} & \val{0.468}{0.010} & \val{0.752}{0.000} & \val{0.772}{0.000} \\
0.40 & \val{0.199}{0.006} & \best{0.139 \pm 0.005} & \val{0.178}{0.002} & \val{0.477}{0.006} & \val{0.538}{0.007} & \val{0.752}{0.000} & \val{0.774}{0.000} \\
0.45 & \best{0.024 \pm 0.007} & \val{0.039}{0.007} & \val{0.058}{0.003} & \val{0.549}{0.004} & \val{0.621}{0.010} & \val{0.749}{0.003} & \val{0.775}{0.000} \\
0.50 & \val{0.180}{0.011} & \val{0.205}{0.005} & \best{0.059 \pm 0.005} & \val{0.631}{0.007} & \val{0.694}{0.006} & \val{0.752}{0.000} & \val{0.775}{0.000} \\
\bottomrule
\end{tabular}
}
\end{table}

\begin{table}[htbp]
\centering
\caption{Brier score on Fashion-MNIST under different noise levels of temporal Fashion-MNIST.}
\label{tab:fashion_mnist_brier}
\renewcommand{\arraystretch}{1.15}
\setlength{\tabcolsep}{5pt}
\arrayrulecolor{black}
\footnotesize
\resizebox{0.7\textwidth}{!}{
\begin{tabular}{c c c c c c c c}
\toprule

\textbf{Noise Level}
& \textbf{GBF}
& \textbf{CVAE}
& \textbf{KALMAN}
& \textbf{SCALE}
& \textbf{DEEPBAYES}
& \textbf{HMMNN}
& \textbf{NN} \\
\midrule
0.0 
& \val{0.045}{0.000}
& \val{0.084}{0.001}
& \val{0.475}{0.000}
& \val{0.049}{0.000}
& \val{0.084}{0.002}
& \best{0.012 \pm 0.000}
& \val{0.040}{0.000} \\

0.1 
& \val{0.047}{0.001}
& \val{0.088}{0.001}
& \val{0.494}{0.000}
& \val{0.051}{0.000}
& \val{0.096}{0.002}
& \best{0.027 \pm 0.002}
& \val{0.092}{0.003} \\

0.2 
& \best{0.050 \pm 0.001}
& \val{0.096}{0.001}
& \val{0.519}{0.000}
& \val{0.059}{0.001}
& \val{0.138}{0.002}
& \val{0.167}{0.010}
& \val{0.495}{0.007} \\

0.3 
& \best{0.062 \pm 0.001}
& \val{0.117}{0.001}
& \val{0.548}{0.000}
& \val{0.084}{0.001}
& \val{0.248}{0.006}
& \val{0.555}{0.021}
& \val{0.935}{0.007} \\

0.4 
& \best{0.093 \pm 0.001}
& \val{0.173}{0.002}
& \val{0.580}{0.000}
& \val{0.172}{0.004}
& \val{0.450}{0.006}
& \val{1.075}{0.057}
& \val{1.224}{0.004} \\

0.5 
& \best{0.193 \pm 0.002}
& \val{0.322}{0.002}
& \val{0.609}{0.000}
& \val{0.438}{0.003}
& \val{0.695}{0.006}
& \val{1.290}{0.041}
& \val{1.379}{0.002} \\
\bottomrule
\end{tabular}
}
\end{table}

\begin{table}[htbp]
\centering
\caption{Expected calibration error on Fashion-MNIST under different noise levels of temporal Fashion-MNIST.}
\label{tab:fashion_mnist_ece}
\renewcommand{\arraystretch}{1.15}
\setlength{\tabcolsep}{5pt}
\arrayrulecolor{black}
\footnotesize
\resizebox{0.7\textwidth}{!}{
\begin{tabular}{c c c c c c c c}
\toprule
\textbf{Noise Level}
& \textbf{GBF}
& \textbf{CVAE}
& \textbf{KALMAN}
& \textbf{SCALE}
& \textbf{DEEPBAYES}
& \textbf{HMMNN}
& \textbf{NN} \\
\midrule
0.0
& \val{0.086}{0.001}
& \val{0.122}{0.001}
& \val{0.467}{0.000}
& \val{0.010}{0.000}
& \val{0.018}{0.002}
& \best{0.006 \pm 0.000}
& \val{0.008}{0.000} \\

0.1
& \val{0.089}{0.002}
& \val{0.123}{0.002}
& \val{0.480}{0.001}
& \best{0.010 \pm 0.001}
& \val{0.015}{0.001}
& \val{0.014}{0.001}
& \val{0.014}{0.003} \\

0.2
& \val{0.097}{0.001}
& \val{0.128}{0.001}
& \val{0.500}{0.001}
& \val{0.011}{0.001}
& \best{0.005 \pm 0.002}
& \val{0.084}{0.005}
& \val{0.217}{0.003} \\

0.3
& \val{0.111}{0.001}
& \val{0.133}{0.001}
& \val{0.515}{0.001}
& \best{0.016 \pm 0.001}
& \val{0.051}{0.002}
& \val{0.278}{0.010}
& \val{0.455}{0.005} \\

0.4
& \val{0.131}{0.002}
& \val{0.131}{0.002}
& \val{0.514}{0.002}
& \best{0.049 \pm 0.003}
& \val{0.152}{0.006}
& \val{0.538}{0.028}
& \val{0.605}{0.002} \\

0.5
& \val{0.118}{0.003}
& \best{0.066 \pm 0.004}
& \val{0.484}{0.003}
& \val{0.164}{0.002}
& \val{0.282}{0.004}
& \val{0.645}{0.020}
& \val{0.686}{0.002} \\
\bottomrule
\end{tabular}
}
\end{table}

\begin{table}[!htbp]
\centering
\caption{Brier score and ECE on the MIT-BIH dataset.}
\label{tab:ecg_calibration}
\renewcommand{\arraystretch}{1.15}
\setlength{\tabcolsep}{12pt}
\arrayrulecolor{black}
\footnotesize
\resizebox{0.45\textwidth}{!}{
\begin{tabular}{lcc}
\toprule
\textbf{Methods} 
& \textbf{Brier Score $\downarrow$} 
& \textbf{ECE $\downarrow$} \\
\midrule
\textbf{GBF} 
& \best{0.330 \pm 0.001} 
& \best{0.093 \pm 0.002} \\

\textbf{CVAE} 
& \val{0.474}{0.001} 
& \val{0.142}{0.002} \\

\textbf{KALMAN} 
& \val{0.575}{0.000} 
& \val{0.257}{0.001} \\

\textbf{SCALE} 
& \val{0.641}{0.002} 
& \val{0.274}{0.001} \\

\textbf{DEEPBAYES} 
& \val{0.623}{0.001} 
& \val{0.244}{0.001} \\

\textbf{HMMNN} 
& \val{0.765}{0.076} 
& \val{0.382}{0.038} \\

\textbf{NN} 
& \val{0.710}{0.003} 
& \val{0.335}{0.002} \\
\bottomrule
\end{tabular}
}
\end{table}

\end{document}